%% file: main.tex
\documentclass{article}

\PassOptionsToPackage{numbers,sort}{natbib}

 \usepackage[preprint]{neurips_2026}

\usepackage[utf8]{inputenc} 
\usepackage[T1]{fontenc}    
\usepackage{hyperref}       
\usepackage{url}            
\usepackage{booktabs}       
\usepackage{amsfonts}       
\usepackage{nicefrac}       
\usepackage{microtype}      
\usepackage{xcolor}         
\usepackage[pdftex]{graphicx}
\usepackage{wrapfig}
\usepackage{tabularx}
\usepackage{longtable}
\usepackage{xltabular}
\usepackage{array}
\usepackage{subcaption}
\usepackage{multirow}
\usepackage{pifont}
\usepackage{xcolor}
\usepackage{alltt}
\usepackage{enumitem}
\usepackage[framemethod=default]{mdframed}

\definecolor{MemoryTokenColor}{RGB}{31,119,180}
\definecolor{OmittedTokenColor}{RGB}{120,120,120}
\definecolor{CheckMarkColor}{RGB}{0,120,0}
\definecolor{CrossMarkColor}{RGB}{180,40,40}
\newcommand{\memorytoken}[1]{\textcolor{MemoryTokenColor}{\texttt{[#1]}}}
\newcommand{\omittedtoken}{\textcolor{OmittedTokenColor}{\texttt{[...]}}}
\newcommand{\cmark}{\textcolor{CheckMarkColor}{\ding{51}}}
\newcommand{\xmark}{\textcolor{CrossMarkColor}{\ding{55}}}
\newcommand{\annline}[1]{\par\hfill\textcolor{OmittedTokenColor}{#1}}
\newmdenv[
  linewidth=0.5pt,
  roundcorner=2pt,
  skipabove=0.5em,
  skipbelow=0.5em,
  innerleftmargin=6pt,
  innerrightmargin=6pt,
  innertopmargin=6pt,
  innerbottommargin=6pt,
  nobreak=true
]{promptbox}
\newcolumntype{Y}{>{\centering\arraybackslash}X}

\title{\textsc{OntoPlan}: An Ontology-Grounded Scene Representation and Agentic Framework for Scalable Robot Task Planning}

\author{%
  Hyeongwoo Nam\quad Woongje Cho\thanks{Equal contribution.}\quad Juwon Kim\footnotemark[1]\quad Jongeun Choi\thanks{Corresponding author.} \\
  School of Mechanical Engineering, Yonsei University, Seoul, Republic of Korea\\
  \texttt{\{namh\_woo, woongje725, 2021145152, jongeunchoi\}@yonsei.ac.kr}
}

\begin{document}

\maketitle

\begin{abstract}
    Large language model (LLM)-based robot task planning is promising for open-ended instruction following, but degrades on long-horizon tasks in large environments. When spatial information is conveyed to the LLM through text, the model can fail to capture spatial context, and token cost grows with environment size. Generating action sequences directly with an LLM also makes it difficult to satisfy the current world state and action preconditions. We address this with an ontology-grounded scene representation that aligns objects, spaces, relations, and states in a shared symbolic vocabulary for spatial reasoning and task planning, and with \textbf{\textsc{OntoPlan}}, an agentic framework that interprets instructions, selectively retrieves task-relevant information, formalizes goals and constraints, and produces executable plans. Across 150 general tasks spanning five indoor environments and three scene scales, \textsc{OntoPlan} achieves 0.89 average task success, compared with 0.27 for the strongest baseline, while using 18.1k total tokens per task on average, about 5.6$\times$ fewer than the most efficient baseline. These advantages persist as scene scale increases, whereas prior methods degrade more sharply in success and remain far more costly in tokens. \textsc{OntoPlan} also responds appropriately to ambiguous or infeasible instructions by asking follow-up questions or reporting insufficient information rather than committing to invalid plans. Code available at \url{https://github.com/namhyeongwoo/OntoPlan}.
\end{abstract}

\input{sections/introduction}

\input{sections/related_work}

\input{sections/ontoplan}

\input{sections/experimental_setup}

\input{sections/results}

\input{sections/conclusion}

\begin{ack}
This work was supported by the National Research Foundation of Korea (NRF) grants funded by the Korea government (MSIT) (Nos.~\mbox{RS-2024-00344732} and \mbox{RS-2026-25505230}). The authors declare no competing interests.
\end{ack}

{
\small
\bibliographystyle{unsrtnat} 
\bibliography{references}
}

\newpage
\input{sections/appendix}

\newpage
\input{checklist.tex}

\end{document}

%% file: sections/introduction.tex
\section{Introduction}
\label{sec:introduction}

Large language models (LLMs) are increasingly used as flexible interfaces for open-ended problem solving~\citep{ruan2023tptu, yang2024swe, jiang2025artificial}. Rather than requiring rigid formal specifications, they let users express intent in natural language and use commonsense knowledge to organize underspecified requests into subgoals or solution steps~\citep{xie2023translating, yang2024embodied, weir2024learning, zhao2023commonsense}. This makes them well suited to robot task planning, where instructions are open-ended and key details are often implicit. Robots must infer the intended task together with the relevant objects, spaces, and constraints.

Robot task planning, however, requires more than language understanding. A valid solution must ground task-relevant objects, spaces, and relations in large scenes and generate long-horizon action sequences that remain executable under the current world state. This exposes two limitations of LLM-only planning. First, flattening structured spatial representations such as 3D scene graphs into text prompts can cause spatial reasoning errors, including hallucinated or omitted relations, while prompt cost grows with scene size~\citep{wang2023can, wu2024can}. Second, directly generated action sequences are unreliable under grounded execution, often violating action preconditions, safety or feasibility constraints, or the intended goal~\citep{valmeekam2023planbench, huang2023grounded}. Alternatively, LLMs can formulate planner-facing PDDL problems for symbolic planners, but this step remains vulnerable to errors such as missing initial facts, misused predicates, and incomplete goals or constraints~\citep{liu2023llm+,liu2025delta}. The central challenge is to maintain consistency across instruction interpretation, scene grounding, and executable planning.

\begin{figure}[t]
    \centering
    \includegraphics[width=\linewidth]{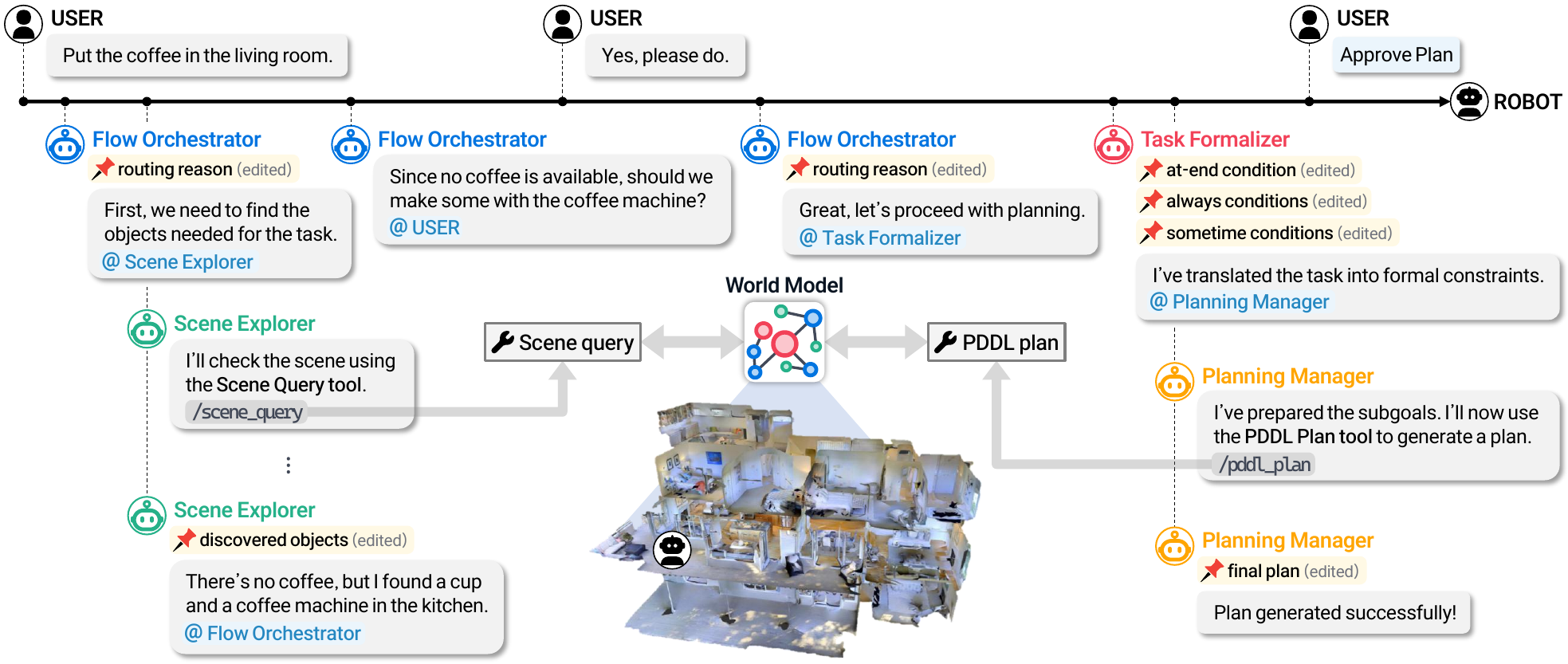}
    \caption{\textbf{Overview of \textsc{OntoPlan}.} From a user instruction, the Flow Orchestrator routes scene grounding, clarification, task formalization, and planning across specialized agents, while the \textit{Scene query} and \textit{PDDL plan} tools operate over a shared ontology-grounded world model.}
    \label{fig:overview}
\end{figure}

We address this challenge with \textbf{\textsc{OntoPlan}}, which separates the roles of LLMs and structured tools. LLM agents handle high-level instruction interpretation, clarification, and goal and constraint formalization, while structured tools handle selective scene access and executable long-horizon planning. To make this separation reliable, \textsc{OntoPlan} maintains a shared symbolic state in an ontology-grounded world model aligned directly with a PDDL domain~\citep{aeronautiques1998pddl}. The \textit{Scene query} tool retrieves task-relevant objects, spaces, and relations from this shared state, while the \textit{PDDL plan} tool generates executable plans from the same world state and explicit constraints. Fig.~\ref{fig:overview} illustrates how the shared world model connects scene grounding, task formalization, planning, and world-state updates.

This paper makes two contributions. First, we propose a PDDL-aligned ontology-grounded scene representation that supports shared scene grounding, direct world-state extraction, and consistent symbolic updates. Second, we develop a tool-using planning framework that combines the \textit{Scene query} tool for selective scene access with the \textit{PDDL plan} tool for executable long-horizon planning. We evaluate \textsc{OntoPlan} across five indoor environments and three scene scales, and further analyze its behavior on ambiguous and infeasible instructions. \textsc{OntoPlan} improves task success and token efficiency over LLM-based baselines and remains more robust as scenes grow or when instructions are ambiguous or infeasible.

%% file: sections/related_work.tex
\section{Related Work}

\noindent \textbf{Structured Scene Grounding.}
Structured scene representations encode objects, spaces, and relations, providing a more reliable substrate for spatial reasoning than free-form scene descriptions~\citep{beetz2018knowrob,armeni20193d,hughes2022hydra,hughes2024foundations,gu2024conceptgraphs,gorlo2026daaam}. SayPlan retrieves task-relevant context from hierarchical 3D scene graphs, passes it to an LLM as text, and replans using feedback from a scene-graph simulator~\citep{rana2023sayplan}. While such methods demonstrate the value of structured grounding, they still route graph structure through natural-language context. Flattening graph-structured information into text can obscure relational dependencies~\citep{wang2023can,fatemi2024talk} and increases prompt length as scenes grow, which often leads to a critical failure in context utilization~\citep{liu2024lost,levy2024same}. This motivates keeping scene knowledge queryable in symbolic form and retrieving only task-relevant evidence~\citep{agia2022taskography,he2024gretriever}.

\noindent \textbf{LLM-Assisted Symbolic Planning.}
Direct LLM planning remains unreliable on formal planning tasks, especially for long-horizon validity and reasoning about state change~\citep{valmeekam2023planbench,valmeekam2023planning,stechly2024chain,li2024embodied}. Hybrid approaches instead use LLMs to interpret instructions or construct planner-facing specifications, then delegate planning to external symbolic planners~\citep{kambhampati2024position, yao2023react}. Prior work has shown that LLMs can translate natural-language goals into formal planning goals~\citep{xie2023translating}, construct or correct PDDL models~\citep{guan2023leveraging,planowl2025,lu2026novelty}, and convert natural-language problems into PDDL~\citep{liu2023llm+, silver2022pddl, gestrin2024nl2plan}. DELTA combines scene-graph pruning with autoregressive subgoal decomposition for embodied long-horizon planning~\citep{liu2025delta}. These approaches reduce reliance on direct plan generation, but remain vulnerable to the formalization errors noted in Section~\ref{sec:introduction}. NL-PDDL extends PDDL into a natural-language representation for lifted regression planning with commonsense entailment in partially observed environments~\citep{liu2026nlpddl}, whereas \textsc{OntoPlan} assumes full observability with a typed vocabulary shared by the world model and PDDL domain.

\noindent \textbf{Interactive Planning under Ambiguity.}
Interactive planning work studies how systems should respond to underspecified instructions or when execution feedback reveals failures~\citep{sun2023adaplanner,shinn2023reflexion}. Inner Monologue incorporates scene descriptions, success signals, and human feedback into closed-loop planning~\citep{huang2022inner}. DEPS revises plans using execution descriptions and self-explanations while selecting among subgoals~\citep{wang2023describe}. Introspective Planning explicitly reasons about uncertainty under ambiguous instructions and asks follow-up questions when needed~\citep{liang2024introspective,ren2023robots}. These studies motivate planning loops that support clarification and feedback-driven revision rather than single-pass plan generation.

\noindent \textbf{Positioning.}
Taken together, prior work spans perceptually grounded construction and querying of spatial and spatio-temporal memories~\citep{hughes2022hydra,honerkamp2024language,gorlo2026daaam} as well as planning over structured scene representations, but does not fully integrate selective access to large structured scenes with executable planning over the same symbolic state. Such memories could serve as upstream inputs after mapping their contents to the ontology--PDDL vocabulary, while \textsc{OntoPlan} uses the resulting symbolic state for scene grounding, task formalization, planner-state extraction, primitive-action synthesis, and symbolic state update.
This architecture instantiates the LLM-modulo framework~\citep{kambhampati2024position} and follows the broader tool-use paradigm for language models~\citep{schick2023toolformer}, supporting consistency across high-level reasoning and grounded execution.

%% file: sections/ontoplan.tex
\begin{figure}[t]
    \centering
    \includegraphics[width=\linewidth]{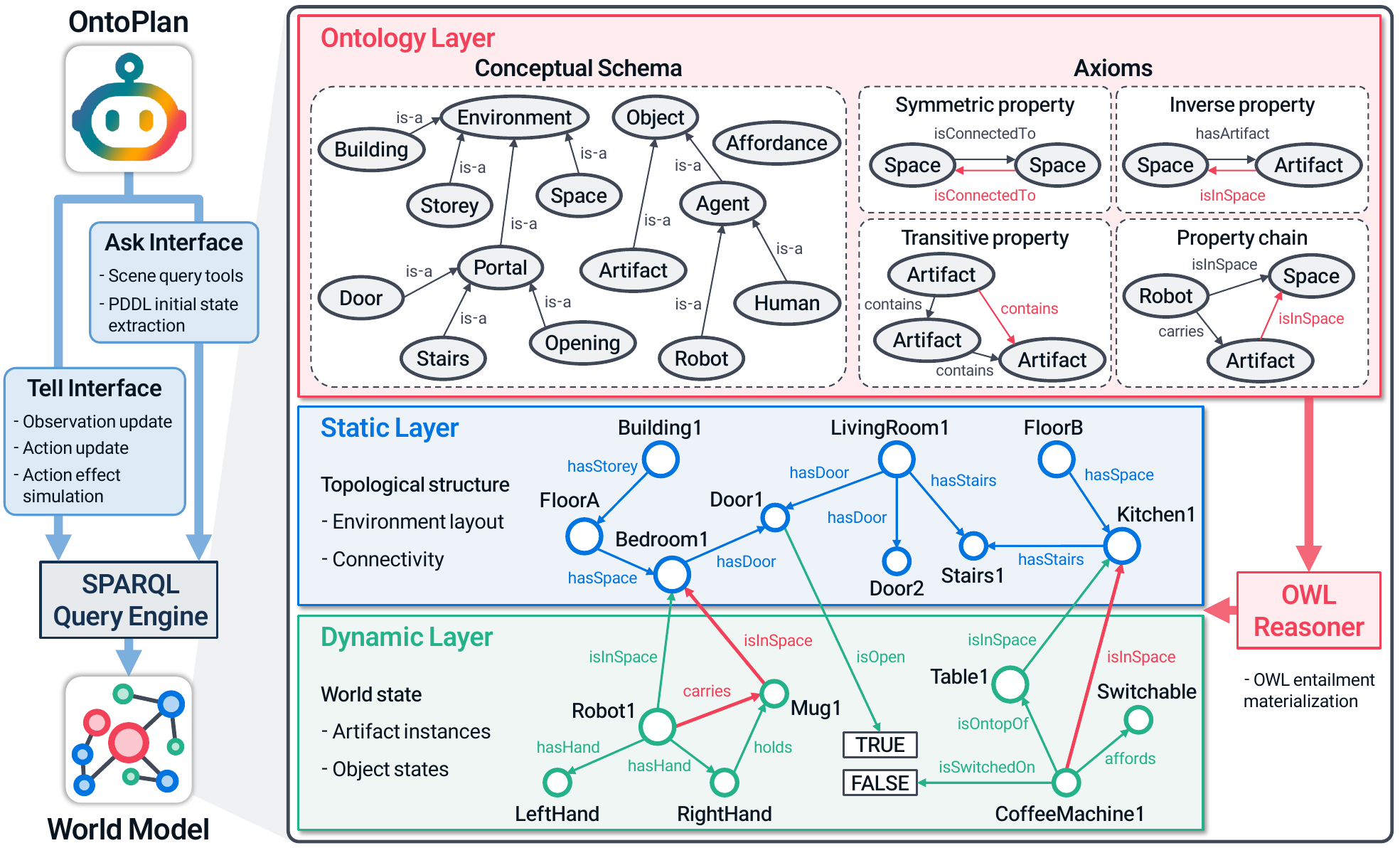}
    \caption{\textbf{World Model.} Ontology-grounded representation with ontology, static, and dynamic layers; \textit{ask} supports scene inspection and planner-state extraction, and \textit{tell} updates dynamic facts from symbolic observations and action effects. Red edges denote facts inferred by the OWL reasoner.}
    \label{fig:world-model}
\end{figure}

\section{\textsc{OntoPlan}}
\textsc{OntoPlan} assumes a fully observable scene represented by symbolic instances and state facts, together with a predefined primitive action set. Its planning loop is organized around a shared ontology-grounded world model aligned with the PDDL domain and two tool interfaces: the \textit{Scene query} tool for scene grounding, which links instruction phrases to scene entities and relations, and the \textit{PDDL plan} tool for generating executable plans. LLM agents coordinate these tools through shared memory while interpreting instructions, requesting clarification, and revising intermediate results.

\subsection{World Model}
\label{sec:world-model}
The world model is the shared symbolic representation for scene understanding, scene grounding, and planning. As shown in Fig.~\ref{fig:world-model}, it consists of three layers: an ontology layer declaring entity types and relation semantics, a static layer storing persistent topology and connectivity, and a dynamic layer storing runtime object instances, states, and spatial relations. Across environments, this fixed schema is instantiated with environment-specific symbolic facts, which in our experiments come from the benchmark scene annotations: topology and connectivity populate the static layer, while object instances, states, and spatial relations populate the dynamic layer. This separates the reusable schema and persistent environment structure from task-time state.

A key design choice is to mirror the ontology directly in the PDDL domain definition: ontology classes map one-to-one to PDDL object types, ontology properties map one-to-one to predicates with the same names, and their associated logical constraints are aligned across the two representations. As a result, scene grounding, task formalization, and planning operate over a shared ontology--PDDL vocabulary, avoiding an intermediate schema-translation stage and keeping grounded context consistent with planner state.

Access to the world model is provided through an ask/tell interface. The \textit{ask} operation executes read-only SPARQL queries for scene inspection and current-state extraction, whereas \textit{tell} updates the dynamic layer by inserting and deleting RDF triples for symbolic observations and action effects. In our implementation, the fully observed scene state is stored as asserted triples. When these triples are loaded or updated, GraphDB applies the declared OWL 2 RL semantics through forward chaining and materializes the resulting entailments. The \textit{ask} operation then queries both asserted facts and materialized inferences. For example, \texttt{carries} is defined by the property chain \texttt{hasHand}$\circ$\texttt{holds}, and an artifact's space is derived through \texttt{isOntopOf}, \texttt{isInsideOf}, and \texttt{isCarriedBy}, so an object carried by the robot is inferred to share the robot's location; in the Marstons Medium scene, materialization expands 2,908 asserted triples to 19,394. The tool interfaces described next are built on this ask/tell mechanism. Appendix~\ref{app:world-model-updates} provides further update and inference examples.

\subsection{System Architecture}

\begin{figure}[t]
    \centering
    \includegraphics[width=\linewidth]{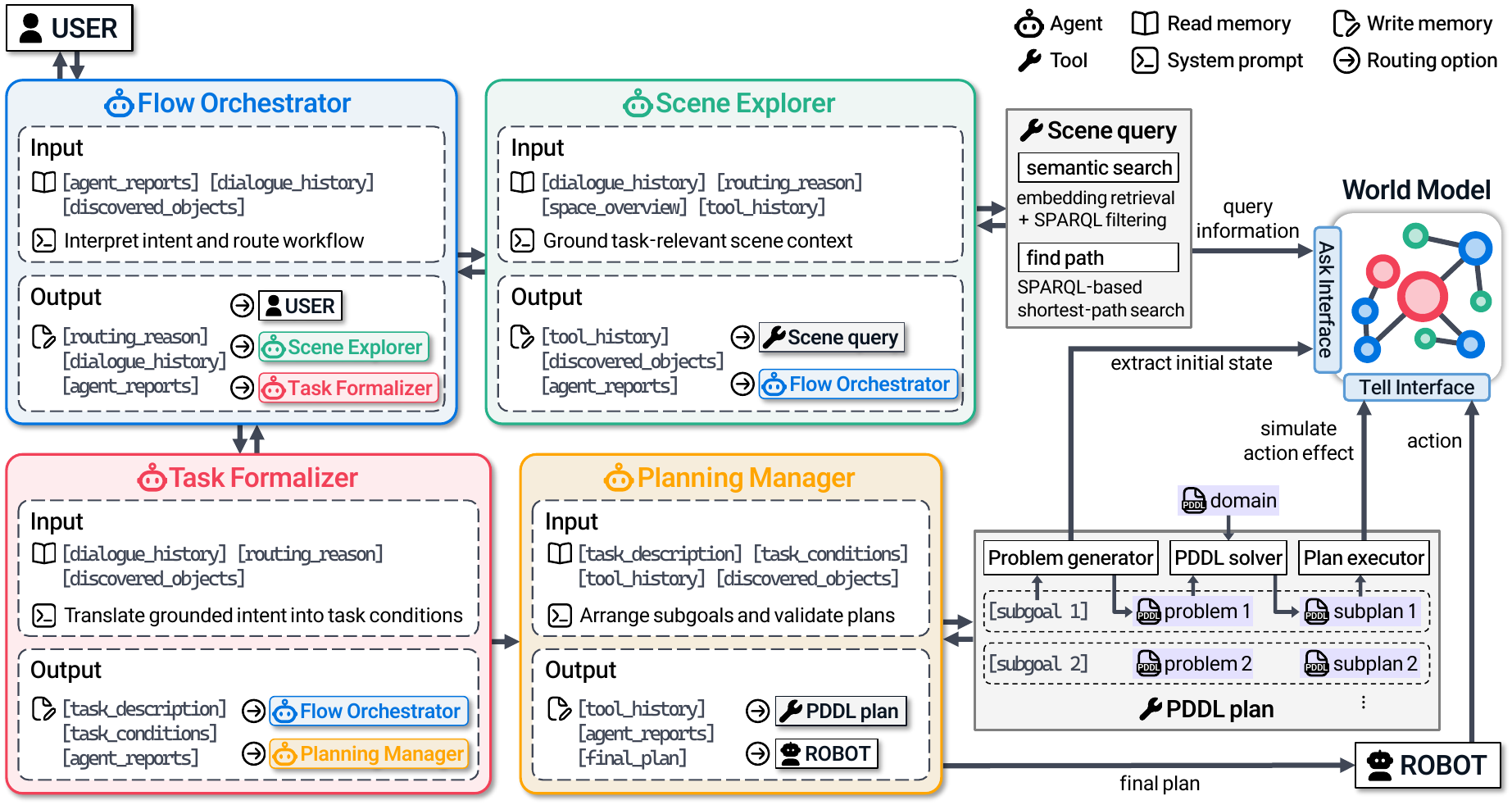}
    \caption{\textbf{System Architecture of \textsc{OntoPlan}.} Flow Orchestrator, Scene Explorer, Task Formalizer, and Planning Manager coordinate through shared memory, while the \textit{Scene query} and \textit{PDDL plan} tools access the shared world model for scene grounding, subgoal planning, and revision. Each agent box lists the memory fields the agent reads (Input) and writes (Output), its role in the system prompt, and its routing options.}
    \label{fig:system-architecture}
\end{figure}

As shown in Fig.~\ref{fig:system-architecture}, \textsc{OntoPlan} is organized around a shared scene representation, two tool interfaces, and shared memory. The \textit{Scene query} tool grounds task-relevant scene evidence, while the \textit{PDDL plan} tool generates executable plans over the current world state. Shared memory stores the current intent interpretation, grounding evidence, unresolved ambiguities, formalized goals and constraints, and planner diagnostics, allowing agents to revise intermediate results as planning proceeds.

\subsubsection{Tool Interfaces}
\textsc{OntoPlan} exposes the world model and planner through two ask/tell-based tool interfaces.
\begin{description}[style=unboxed, leftmargin=0pt, parsep=0pt, itemsep=4pt, topsep=4pt]
    \item[\textbf{Scene query.}] The \textit{Scene query} tool inspects graph-structured knowledge in the world model through two functions. \texttt{semantic\_search} combines embedding-based retrieval with SPARQL filtering to identify task-relevant objects and spaces, while \texttt{find\_path} computes shortest paths over the scene connectivity graph using SPARQL. During scene grounding, the agent issues task-dependent queries, records the returned evidence in shared memory, and refines follow-up queries as needed. When restrictive filters return no match, \texttt{semantic\_search} can broaden the query scope and return hint fields with similar categories. Scene grounding thus proceeds as iterative evidence gathering over the symbolic scene representation, rather than by serializing the full scene graph into the prompt.
    \item[\textbf{PDDL plan.}] The \textit{PDDL plan} tool synthesizes or repairs long-horizon executable plans from a formalized task specification. It extracts the current world state from the world model, compiles trajectory-level \texttt{always} conditions into the planning model, and solves ordered subgoals sequentially. Each planning problem is built with SPARQL from the artifacts named in the subgoal, the containers, surfaces, and contents linked to them, the full navigation topology, and the robot state, so unrelated objects do not enlarge it. Because Fast Downward~\citep{helmert2006fast}, the planner in our implementation, does not support PDDL 3 trajectory constraints, each \texttt{always} condition is compiled into the domain as a violation flag that any action falsifying the condition sets, and every problem requires the flag to remain false. After each successful subgoal, the tool simulates action effects, writes the resulting updates back to the world model, and uses the updated state for the next subgoal, keeping the parts of the final goal already achieved as goals so that later subgoals do not undo them. The tool returns either a plan or failure diagnostics, so the agent workflow does not need to construct a full PDDL problem directly. Fast Downward handles action-level validity and state transitions instead of relying on the LLM to generate executable action sequences directly.
\end{description}

\subsubsection{Agent Workflow}
On top of these tools, \textsc{OntoPlan} factorizes the planning loop into four roles with distinct responsibilities: workflow control, scene grounding, task formalization, and executable planning/revision. This decomposition narrows each agent's reasoning scope while preserving a shared symbolic task context. Detailed prompts are provided in Appendix~\ref{app:agent-prompts}.
\begin{description}[style=unboxed, leftmargin=0pt, parsep=0pt, itemsep=4pt, topsep=4pt]
    \item[\textbf{Flow Orchestrator.}] The Flow Orchestrator maintains the current intent interpretation from the instruction, dialogue history, and shared memory. It routes the workflow by deciding whether to ask the user for clarification, continue scene grounding, or proceed to task formalization.
    \item[\textbf{Scene Explorer.}] The Scene Explorer uses the \textit{Scene query} tool to retrieve task-relevant objects, spaces, relations, and states from the world model~\citep{yao2023react}. It records the evidence and candidate entities in shared memory, then returns control once the task is sufficiently grounded or when missing or ambiguous facts require further routing.
    \item[\textbf{Task Formalizer.}] The Task Formalizer converts the interpreted intent and grounded scene context into planner-facing final-state goals and trajectory constraints. If the grounding evidence is sufficient, it writes the formalized conditions to shared memory and hands them to the Planning Manager; otherwise, it requests further grounding.
    \item[\textbf{Planning Manager.}] The Planning Manager organizes the formalized goals and constraints into ordered subgoals, invokes the \textit{PDDL plan} tool on the current world state, and interprets the resulting plan or failure diagnostics. When planning fails, it uses this feedback to revise the subgoal structure and trigger replanning.
\end{description}

%% file: sections/experimental_setup.tex
\section{Experimental Setup}
\label{sec:experimental-setup}
We evaluate \textsc{OntoPlan} on long-horizon robot task planning in large-scale indoor environments. Our evaluation uses five indoor environments, three scene scales, 150 general instructions, and two diagnostic subsets (ambiguous and infeasible). We compare against SayPlan and DELTA under shared LLM and planner settings using planning reliability, task success, and token-usage metrics.

\subsection{Datasets}
\noindent
\begin{minipage}[t]{0.43\linewidth}
\vspace{0pt}
\paragraph{Environments.} The environment dataset consists of five indoor spaces from the 3D Scene Graph dataset~\citep{armeni20193d}: Klickitat, Lakeville, Lindenwood, Marstons, and Muleshoe. We deterministically map the original annotations into a fixed ontology schema to form the \emph{Small} split, manually augment the resulting scenes to form the \emph{Medium} split, and derive the \emph{Large} split from \emph{Medium} through constrained random augmentation to triple the object count. All scales share the same ontology and PDDL schema. Table~\ref{tab:dataset-stats} reports the structural statistics and object counts, and Appendix~\ref{app:dataset-scale-augmentation} visualizes the scene graphs across scales.
\end{minipage}\hfill
\begin{minipage}[t]{0.54\linewidth}
\vspace{0pt}
\centering
\small
\setlength{\tabcolsep}{3pt}
\captionsetup{hypcap=false}
\captionof{table}{\textbf{Evaluation Environment Statistics.} Floors, rooms, and object counts for the five environments at Small, Medium, and Large scales.}
\label{tab:dataset-stats}
\begin{tabular*}{\linewidth}{@{\extracolsep{\fill}}lccccc}
\toprule
Name & Floors & Rooms & \multicolumn{3}{c}{Objects} \\
\cmidrule(r){4-6}
 &  &  & Small & Medium & Large \\
\midrule
Klickitat & 3 & 28 & 84 & 385 & 1155 \\
Lakeville & 2 & 19 & 83 & 387 & 1161 \\
Lindenwood & 2 & 36 & 59 & 318 & 954 \\
Marstons & 4 & 31 & 50 & 388 & 1164 \\
Muleshoe & 2 & 24 & 121 & 423 & 1269 \\
\bottomrule
\end{tabular*}
\end{minipage}

\noindent
\begin{minipage}[t]{0.39\linewidth}
\vspace{0pt}
\paragraph{Task Instructions.} For each environment and scale, we manually wrote 10 general task instructions, yielding 150 general tasks in total. We also constructed two diagnostic subsets with 10 tasks each: \textit{Ambiguous} and \textit{Infeasible}. The general tasks form the primary evaluation set, while the diagnostic subsets are used to analyze behavior under underspecified or unsatisfiable requests rather than to support standalone statistical claims. Representative examples are shown in Table~\ref{tab:task-instruction-dataset}, and additional examples are provided in Appendix~\ref{app:general-instruction-examples}. Each task is paired with evaluation checks used for scoring: \texttt{at\_end}, \texttt{always}, and \texttt{sometime}, which specify the final goal, trajectory-wide constraints, and intermediate states that must be achieved at least once, and, for six tasks, \texttt{ordering}, which requires one state to first hold no later than another.
\end{minipage}\hfill
\begin{minipage}[t]{0.58\linewidth}
\vspace{0pt}
\centering
\small
\setlength{\tabcolsep}{3pt}
\captionsetup{hypcap=false}
\captionof{table}{\textbf{Representative Task Instructions.} Abridged examples from the three subsets, with gray annotations summarizing their evaluation context.}
\label{tab:task-instruction-dataset}
\begingroup
\renewcommand{\annline}[1]{\par\hfill{\footnotesize\textcolor{OmittedTokenColor}{#1}}}
\begin{tabularx}{\linewidth}{@{}lX@{}}
\toprule
Type & Examples \\
\midrule
\multirow{2}{*}{General} & I want to ventilate the house. Open all closed doors, except the home office door.
\annline{$\rightarrow$ (always) keep home office door closed.}
\annline{$\rightarrow$ (at\_end) all doors except home office door are open.} \\
\cmidrule(lr){2-2}
& Wash Sam's T-shirt from the bed on Floor A in the washer, then put it back on the bed.
\annline{$\rightarrow$ (sometime) T-shirt in washer while washer is on.}
\annline{$\rightarrow$ (at\_end) T-shirt is back on the bed.} \\
\midrule
Ambiguous & Bring the book I was reading yesterday from the living room to my room.
\annline{$\rightarrow$ (ambiguity) book and room unspecified.} \\
\midrule
Infeasible & I want to skateboard again after a long time. Find the skateboard in the living room on Floor B and bring it to the lobby.
\annline{$\rightarrow$ (infeasibility) no skateboard in the environment.} \\
\bottomrule
\end{tabularx}
\endgroup
\end{minipage}

\subsection{Evaluation Metrics}
\label{sec:evaluation-metrics}

The evaluation checks are used only for scoring; neither \textsc{OntoPlan} nor the baselines receive them as input. \textit{Plan Returned} measures whether the system returns a symbolic plan for the task. Each returned action sequence is then symbolically executed from the current state until either all actions are applied or the first precondition failure occurs; Appendix~\ref{app:experimental-protocol-details} provides the full symbolic evaluation procedure. \textit{Plan Executable} measures whether the full sequence completes without failure. \textit{Goal Reached} measures whether the reached state satisfies the \texttt{at\_end} condition; because this state may result from an executable prefix, Goal Reached can exceed Plan Executable. \textit{Task Success} requires full execution and satisfaction of all applicable checks, including \texttt{at\_end}, \texttt{always}, \texttt{sometime}, and \texttt{ordering}.

Efficiency is measured using \textit{Total Tokens per Task}, \textit{Input Tokens per Call}, and, in the ablation, \textit{Peak Input Tokens per Call}. The first sums input and output tokens over all LLM calls for a task.\footnote{This includes calls in runs that terminated after exceeding the model's context limit, which occurs only for SayPlan.} The second averages input tokens over those calls. The third is the largest single-call input per task.

\subsection{Implementation Details}
\label{sec:implementation-details}

We compare \textsc{OntoPlan} with SayPlan and DELTA under matched task instances, initial states, LLM settings, and success checks. We use the released DELTA codebase and implement SayPlan following its original hierarchical scene-graph retrieval pipeline. All methods use GPT-4o with temperature 0~\citep{hurst2024gpt}, share the same 16 primitive actions, and are evaluated with the same action preconditions, effects, and checks. DELTA and \textsc{OntoPlan} start from the same PDDL domain and use the same Fast Downward configuration~\citep{helmert2006fast} (lazy weighted A* with the FF heuristic, $w=2$), and SayPlan and \textsc{OntoPlan} may use up to 20 scene-exploration LLM calls and 5 plan-generation calls. Appendix~\ref{app:experimental-protocol-details} gives further details.

The main quantitative comparison uses a non-interactive protocol. Because SayPlan and DELTA do not support user interaction during planning, \textsc{OntoPlan} also runs with a fixed automatic reply to clarification requests in this setting: ``\textit{Do not ask follow-up questions. Make reasonable assumptions and continue planning.}'' For the two diagnostic subsets, we report additional analyses: for the ambiguous subset, a user-interactive setting in which a human answers \textsc{OntoPlan}'s clarification questions; and for the infeasible subset, the initial response types to unsatisfiable requests.

%% file: sections/results.tex
\section{Results}

We organize the results around five questions: how \textsc{OntoPlan} compares with baselines on general-task success and token efficiency, how it behaves on ambiguous and infeasible instructions, how performance and cost scale with scene size, how the two tool-enabled paths relate to task reliability and scalability, and how the full workflow behaves in qualitative multi-turn examples.

\subsection{Performance on General Tasks}
Table~\ref{tab:robot-task-planning} shows that \textsc{OntoPlan} consistently outperforms both baselines across all five environments, reaching 0.89 average task success versus 0.27 for SayPlan and 0.13 for DELTA while using 18.1k total tokens per task versus 216.0k and 100.6k, respectively. Its smaller gap between Plan Returned and Plan Executable indicates that its returned plans more often remain executable under the evaluated world state, while the lower token cost reflects selective scene querying. Appendix~\ref{app:statistical-significance} provides paired 95\% bootstrap confidence intervals, stratified by scene scale, for the four binary performance metrics in Table~\ref{tab:robot-task-planning}. All 149 plans returned by \textsc{OntoPlan} execute without precondition failures; its remaining failures arise before planning, mostly from grounding a referring expression to the wrong object or place (Appendix~\ref{app:failure-analysis}). Baseline plans go unreturned mainly because of incomplete specifications or invalid search commands, not timeouts (Appendix~\ref{app:unreturned-plans}).

\begin{table}[t]
\centering
\small
\setlength{\tabcolsep}{4pt}
\caption{\textbf{General-Task Performance by Environment.} Results are averaged over the three scene scales for each environment. All four performance metrics are rates, and token metrics are reported in $\times 10^3$. Bold marks the best value in each environment.}
\label{tab:robot-task-planning}
\begin{tabularx}{\linewidth}{@{}ll*{6}{Y}@{}}
\toprule
Environment & Method & \shortstack{Plan\\Returned} & \shortstack{Plan\\Executable} &
\shortstack{Goal\\Reached} & \shortstack{Task\\Success} &
\shortstack{Input Tokens\\per Call} & \shortstack{Total Tokens\\per Task} \\
\midrule
\multirow{3}{*}{Klickitat} & SayPlan & 0.67 & 0.47 & 0.33 & 0.27 & 30.9 & 221.7 \\
 & DELTA & 0.63 & 0.23 & 0.23 & 0.20 & 23.9 & 101.3 \\
 & \textsc{OntoPlan} & \textbf{1.00} & \textbf{1.00} & \textbf{0.93} & \textbf{0.90} & \textbf{2.6} & \textbf{17.3} \\
\midrule
\multirow{3}{*}{Lakeville} & SayPlan & 0.57 & 0.40 & 0.33 & 0.33 & 33.3 & 205.5 \\
 & DELTA & 0.47 & 0.23 & 0.17 & 0.17 & 23.4 & 98.8 \\
 & \textsc{OntoPlan} & \textbf{1.00} & \textbf{1.00} & \textbf{0.97} & \textbf{0.97} & \textbf{2.6} & \textbf{18.8} \\
\midrule
\multirow{3}{*}{Lindenwood} & SayPlan & 0.63 & 0.47 & 0.40 & 0.40 & 34.5 & 206.6 \\
 & DELTA & 0.37 & 0.17 & 0.23 & 0.17 & 23.4 & 98.9 \\
 & \textsc{OntoPlan} & \textbf{1.00} & \textbf{1.00} & \textbf{0.87} & \textbf{0.87} & \textbf{2.6} & \textbf{17.5} \\
\midrule
\multirow{3}{*}{Marstons} & SayPlan & 0.73 & 0.43 & 0.30 & 0.27 & 37.1 & 255.6 \\
 & DELTA & 0.27 & 0.17 & 0.17 & 0.10 & 23.9 & 101.3 \\
 & \textsc{OntoPlan} & \textbf{1.00} & \textbf{1.00} & \textbf{0.77} & \textbf{0.77} & \textbf{2.6} & \textbf{18.7} \\
\midrule
\multirow{3}{*}{Muleshoe} & SayPlan & 0.70 & 0.17 & 0.20 & 0.10 & 31.9 & 190.4 \\
 & DELTA & 0.47 & 0.03 & 0.03 & 0.03 & 24.2 & 102.7 \\
 & \textsc{OntoPlan} & \textbf{0.97} & \textbf{0.97} & \textbf{0.93} & \textbf{0.93} & \textbf{2.6} & \textbf{18.1} \\
\bottomrule
\end{tabularx}
\end{table}

\subsection{Handling Ambiguous and Infeasible Instructions}

Although the ambiguous and infeasible subsets each contain only 10 tasks, they reveal behavioral differences not captured by aggregate success rates. Table~\ref{tab:ambiguous-instructions} shows that baselines recover poorly from underspecified requests, whereas \textsc{OntoPlan} uses clarification before planning; with user answers instead of the fixed automatic reply from Section~\ref{sec:implementation-details}, it reaches the goal in 8/10 tasks instead of 1/10. For infeasible tasks, Table~\ref{tab:infeasible-behavior} groups responses into four categories: \textit{Clarification Request}, when the system identifies missing, conflicting, or infeasible grounding evidence and asks for clarification or an alternative; \textit{Misdirected Plan}, when the plan targets the wrong object or goal; \textit{Inexecutable Plan}, when symbolic execution fails; and \textit{Planning Failure}, when no plan is returned. Because SayPlan and DELTA cannot request clarification in our protocol, they mostly fall into less desirable categories, especially planning failure. In contrast, \textsc{OntoPlan} most often asks for clarification rather than committing to an invalid plan. Full per-instruction results are provided in Appendices~\ref{app:ambiguous-full-results} and~\ref{app:infeasible-full-results}.

\begin{table*}[t]
\centering

\begin{minipage}[t]{0.49\textwidth}
\centering
\footnotesize
\setlength{\tabcolsep}{1.5pt}
\setlength{\belowcaptionskip}{0.4em}
\captionof{table}{\textbf{Performance on Ambiguous Instructions.} Results on 10 ambiguous tasks, reported as successful cases. Auto replies use the fixed reply of Section~\ref{sec:implementation-details}; user replies answer the clarification questions directly.}
\label{tab:ambiguous-instructions}
\begin{tabularx}{\linewidth}{@{}lYYY@{}}
\toprule
Method & \shortstack{Plan\\Returned} & \shortstack{Plan\\Executable} & \shortstack{Goal\\Reached\rlap{$^\dagger$}} \\
\midrule
SayPlan & 5/10 & 4/10 & 0/10 \\
DELTA & 5/10 & 3/10 & 0/10 \\
\textsc{OntoPlan} {\scriptsize (auto replies)} & 8/10 & 8/10 & 1/10 \\
\cmidrule{1-4}
\textsc{OntoPlan} {\scriptsize (user replies)} & \textbf{10/10} & \textbf{10/10} & \textbf{8/10} \\
\bottomrule
\end{tabularx}
\vspace{0.2em}
{\raggedright\footnotesize $^\dagger$ Ambiguous tasks use only \texttt{at\_end}; Task Success is omitted.\par}
\end{minipage}\hfill
\begin{minipage}[t]{0.49\textwidth}
\centering
\footnotesize
\setlength{\tabcolsep}{2pt}
\setlength{\belowcaptionskip}{0.4em}
\captionof{table}{\textbf{Responses to Infeasible Instructions.} Results on 10 infeasible tasks, reported as response counts by category.}
\label{tab:infeasible-behavior}
\begin{tabularx}{\linewidth}{@{}>{\raggedright\arraybackslash}Xcccc@{}}
\toprule
Method & \shortstack{Clarification\\Request\rlap{$^\dagger$}} & \shortstack{Misdirected\\Plan} & \shortstack{Inexecutable\\Plan} & \shortstack{Planning\\Failure} \\
\midrule
SayPlan & --- & 3/10 & 0/10 & 7/10 \\
DELTA & --- & 1/10 & 1/10 & 8/10 \\
\cmidrule{1-5}
\textsc{OntoPlan} & \textbf{9/10} & 1/10 & 0/10 & 0/10 \\
\bottomrule
\end{tabularx}
\vspace{0.2em}
{\raggedright\footnotesize $^\dagger$ Not supported for SayPlan or DELTA.\par}
\end{minipage}

\end{table*}

\subsection{Scalability Analysis}
\label{sec:scalability-analysis}

\begin{figure}[t]
\centering
\begin{subfigure}[t]{0.49\linewidth}
\centering
\includegraphics[width=\linewidth]{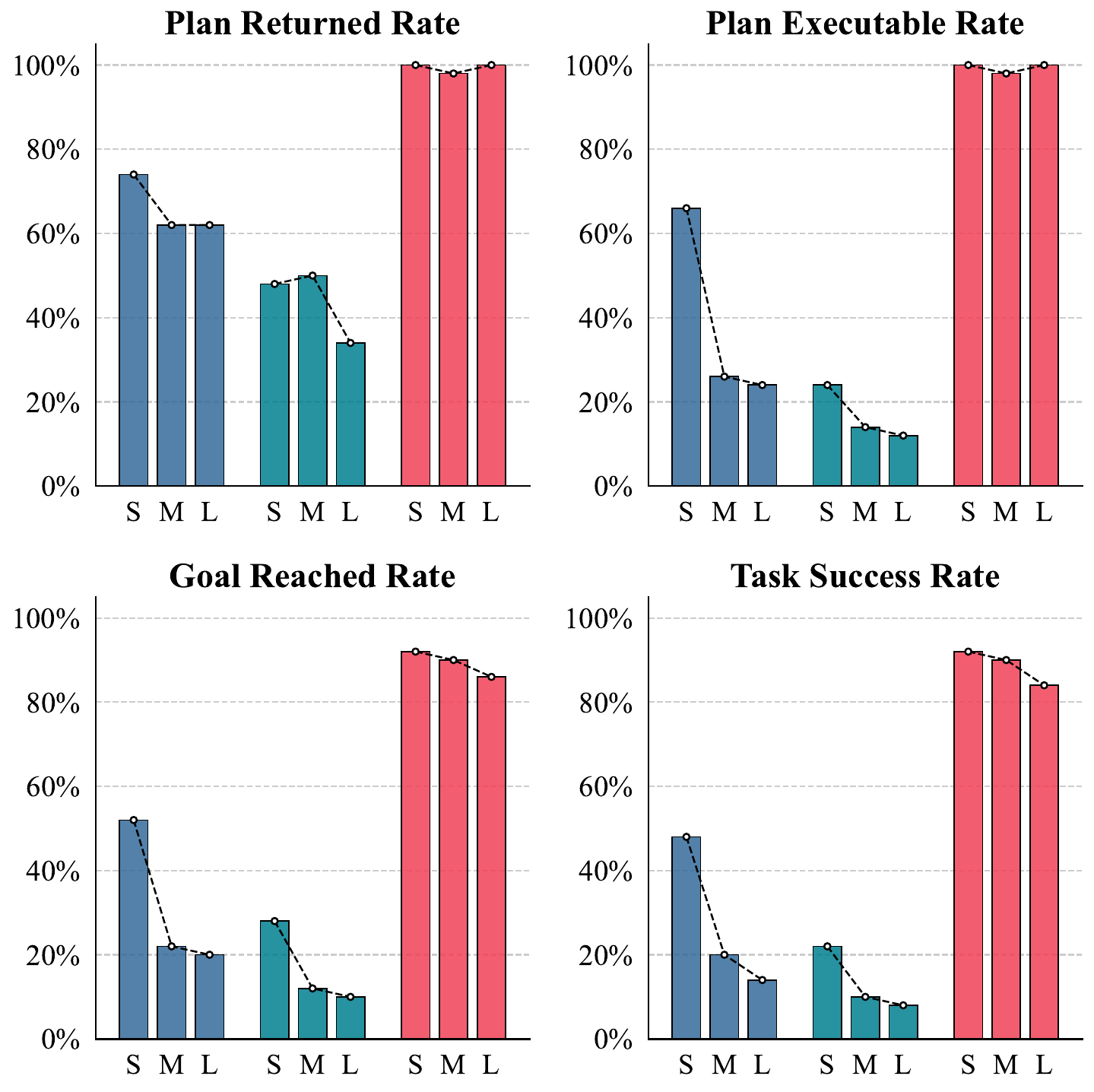}
\caption{Bar plots of success metrics by scene scale.}
\label{fig:scalability-success-barplot}
\end{subfigure}\hfill
\begin{subfigure}[t]{0.49\linewidth}
\centering
\includegraphics[width=\linewidth]{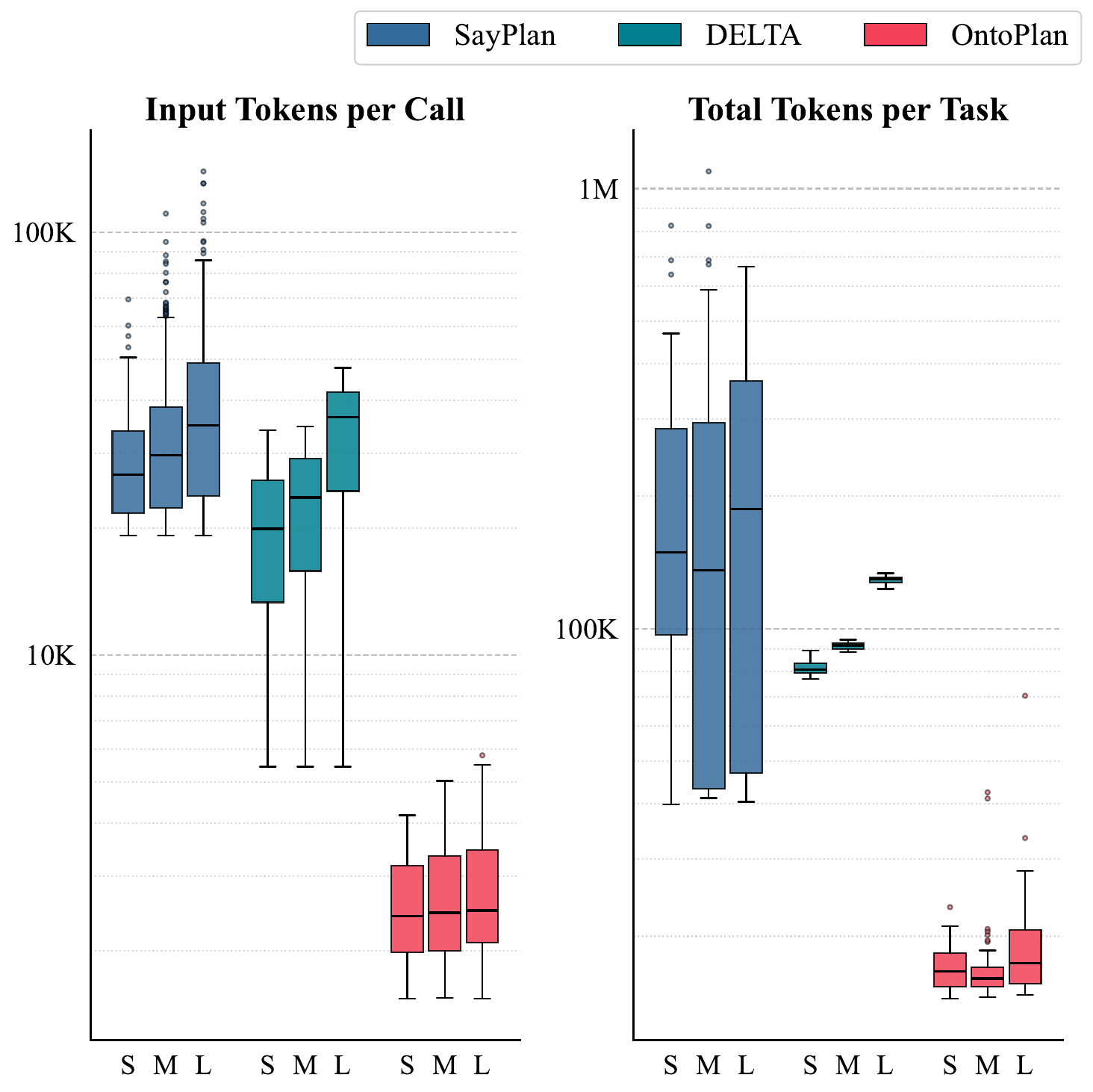}
\caption{Box plots of input tokens per call and total tokens per task by scene scale (log scale).}
\label{fig:scalability-token-boxplot}
\end{subfigure}
\caption{\textbf{Scalability Across Scene Scales.} \textsc{OntoPlan} degrades more gracefully than the baselines while remaining more token-efficient as scene complexity increases. Each scale comprises 50 tasks, and dashed lines connect a method's rates across scales.}
\label{fig:scalability-plots}
\end{figure}

Figure~\ref{fig:scalability-plots}(a)--(b) shows that \textsc{OntoPlan} maintains the highest success rates across all four metrics as scene complexity increases from Small to Medium to Large, while its input and total token usage remain comparatively flat across scales. By contrast, SayPlan shows larger variance and heavier token tails, and DELTA remains substantially more token-intensive than \textsc{OntoPlan} at larger scales. Taken together, the plots indicate that \textsc{OntoPlan} degrades more gracefully with scene size while keeping its token cost low and nearly flat.

\subsection{Ablation Study}
\label{sec:ablation-study}

Table~\ref{tab:tool-ablation} compares the full system with two interface-level variants under matched task and evaluation settings. Without the \textit{Scene query} tool, the full scene is serialized into the prompt, so the LLM must identify task-relevant information from the entire scene description. The Large-to-Small ratio consequently increases from 1.15$\times$ to 4.02$\times$ for \textit{Total Tokens per Task} and from 1.06$\times$ to 8.13$\times$ for \textit{Peak Input Tokens per Call}, indicating that selective retrieval is central to the observed token scalability, while task success remains statistically indistinguishable from that of the full system (Appendix~\ref{app:statistical-significance}). In the w/o \textit{PDDL plan} variant, the \textit{Scene query} tool is retained, but the integrated planning path---including ordered-subgoal planning, Fast Downward search, symbolic state-transition handling, planner diagnostics, and feedback-based replanning---is replaced by a single LLM call that receives the filtered scene context, the formalized task conditions, and the same PDDL domain, and generates the complete primitive-action sequence without the ordered-subgoal loop or planner-diagnostic-based revision. Task success decreases from 0.89 to 0.31, reflecting the integrated planning path as a whole rather than Fast Downward search or ontology--PDDL alignment in isolation. Overall, the results associate selective scene access with scalability and the integrated planning path with task reliability.

\begin{table*}[t]
\centering
\footnotesize
\setlength{\tabcolsep}{2pt}
\caption{\textbf{Ablation of \textit{Scene query} and \textit{PDDL plan} across scene scales.} The w/o \textit{Scene query} and w/o \textit{PDDL plan} variants replace selective retrieval with full-scene serialization and the integrated planning path with direct LLM action-sequence generation, respectively. Avg. denotes the mean across scales; token metrics are reported in $\times 10^3$, and L/S denotes the Large-to-Small ratio.}
\label{tab:tool-ablation}
\begin{tabularx}{\textwidth}{@{}>{\raggedright\arraybackslash}p{0.16\textwidth}*{12}{Y}@{}}
\toprule
\multirow{2}{*}{Setting} & \multicolumn{4}{c}{Task Success} & \multicolumn{4}{c}{Total Tokens per Task} & \multicolumn{4}{c}{Peak Input Tokens per Call} \\
\cmidrule(lr){2-5}\cmidrule(lr){6-9}\cmidrule(lr){10-13}
& S & M & L & Avg. & S & M & L & L/S & S & M & L & L/S \\
\midrule
Full system & \textbf{0.92} & \textbf{0.90} & \textbf{0.84} & \textbf{0.89} & \textbf{17.1} & \textbf{17.5} & \textbf{19.7} & $1.15\times$ & \textbf{3.6} & \textbf{3.7} & \textbf{3.9} & $1.06\times$ \\
\quad w/o Scene query & 0.86 & \textbf{0.90} & \textbf{0.84} & 0.87 & 18.1 & 32.5 & 72.7 & $4.02\times$ & 7.6 & 22.3 & 62.2 & $8.13\times$ \\
\quad w/o PDDL plan & 0.32 & 0.26 & 0.36 & 0.31 & 23.6 & 25.0 & 26.9 & \textbf{\boldmath$1.14\times$} & 8.4 & 8.5 & 8.4 & \textbf{\boldmath$1.01\times$} \\
\bottomrule
\end{tabularx}
\end{table*}

\subsection{Qualitative Analysis}

\begin{figure}[t]
\centering
\includegraphics[width=\linewidth]{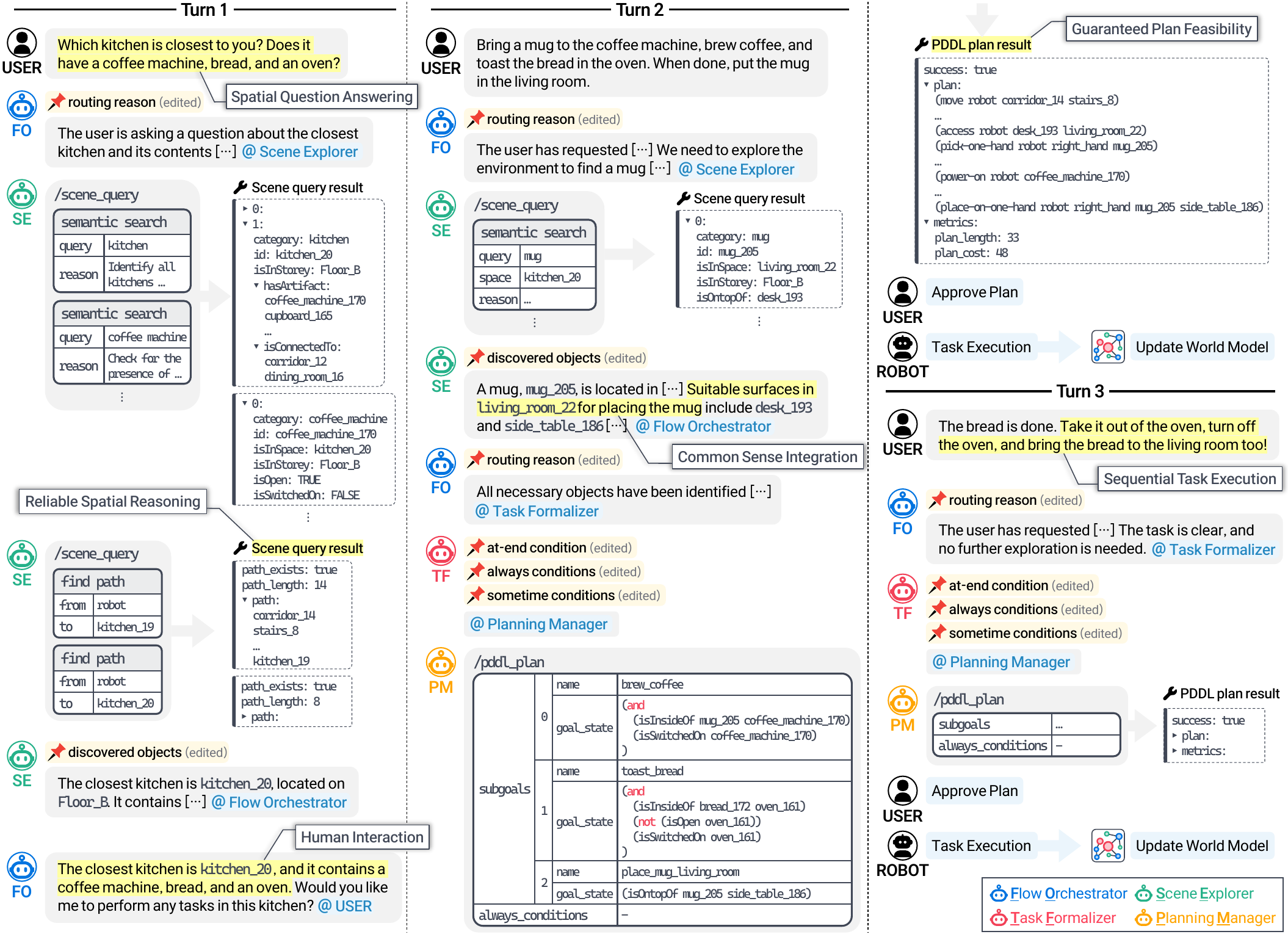}
\caption{\textbf{Qualitative Examples of \textsc{OntoPlan}.} A trace of spatial question answering, commonsense task grounding, executable planning, and sequential execution with world model updates.}
\label{fig:qualitative-results}
\end{figure}

Figure~\ref{fig:qualitative-results} illustrates how \textsc{OntoPlan} uses a shared world model across question answering, task grounding, executable planning, and follow-up execution. The system retrieves only task-relevant scene facts, translates them into planner-ready goals and constraints, and invokes symbolic planning over the current state. The trace highlights three capabilities not captured by aggregate metrics: path-based spatial question answering, grounding of implicit placement choices such as suitable support surfaces, and follow-up planning after action effects update the world model. Together, they show the two tools interacting over a shared symbolic state across user turns.

%% file: sections/conclusion.tex
\section{Conclusion}
\label{sec:conclusion}

\textsc{OntoPlan} shows that scalable LLM-based robot task planning benefits from delegating structured scene access and executable plan synthesis to dedicated tools over a shared symbolic representation. Its ontology-grounded world model aligns directly with the PDDL domain, allowing scene grounding, task formalization, state extraction, and action-effect updates to use the same symbolic vocabulary. Across the evaluated environments, \textsc{OntoPlan} improves task success and token efficiency relative to representative baselines while better handling ambiguity and infeasibility.

The current study assumes a fully instantiated, fully observable symbolic world model and does not evaluate raw-sensor perception, learned scene reconstruction, or partial observability, so the results should be interpreted as evidence for planning quality rather than end-to-end real-world autonomy. Extending \textsc{OntoPlan} to partial observability will require perception and state-estimation modules~\citep{hughes2022hydra,gorlo2026daaam} that convert sensor observations into asserted symbolic updates. The ontology can enrich these asserted facts through declared entailments, but does not recover unobserved state by itself. The ontology--PDDL alignment also adds a modeling cost. Runtime state changes modify only dynamic facts, and new environments expressible under the existing schema and action set require only new static and instance facts, whereas genuinely new semantics or robot capabilities require coordinated extensions to both the ontology and the PDDL domain. Automatic cross-domain transfer, schema and action-model induction, and integration with vision-language-action models and real robot systems remain future work.

%% file: sections/appendix.tex
\appendix
\section{World Model Update and OWL Inference Examples}
\label{app:world-model-updates}

This appendix provides examples of the ask/tell update mechanism and OWL inference behavior described in Section~\ref{sec:world-model}. In the evaluated setting, the asserted graph represents the fully observed scene state. When triples are loaded or updated, GraphDB applies the declared OWL 2 RL semantics through forward chaining and materializes entailments that subsequent queries read together with the asserted facts. Figure~\ref{fig:wm-viz-comparison} shows the world model before and after OWL reasoning in the Marstons Medium scene, and Table~\ref{tab:plan-diffs} lists representative direct updates together with selected inferred facts and the axiom patterns that support those inferences under actions and symbolic observations.

\begin{center}
\begin{minipage}{\textwidth}
\centering
\captionsetup{type=figure}
\begin{subfigure}[t]{0.49\textwidth}
\centering
\includegraphics[width=0.714\linewidth]{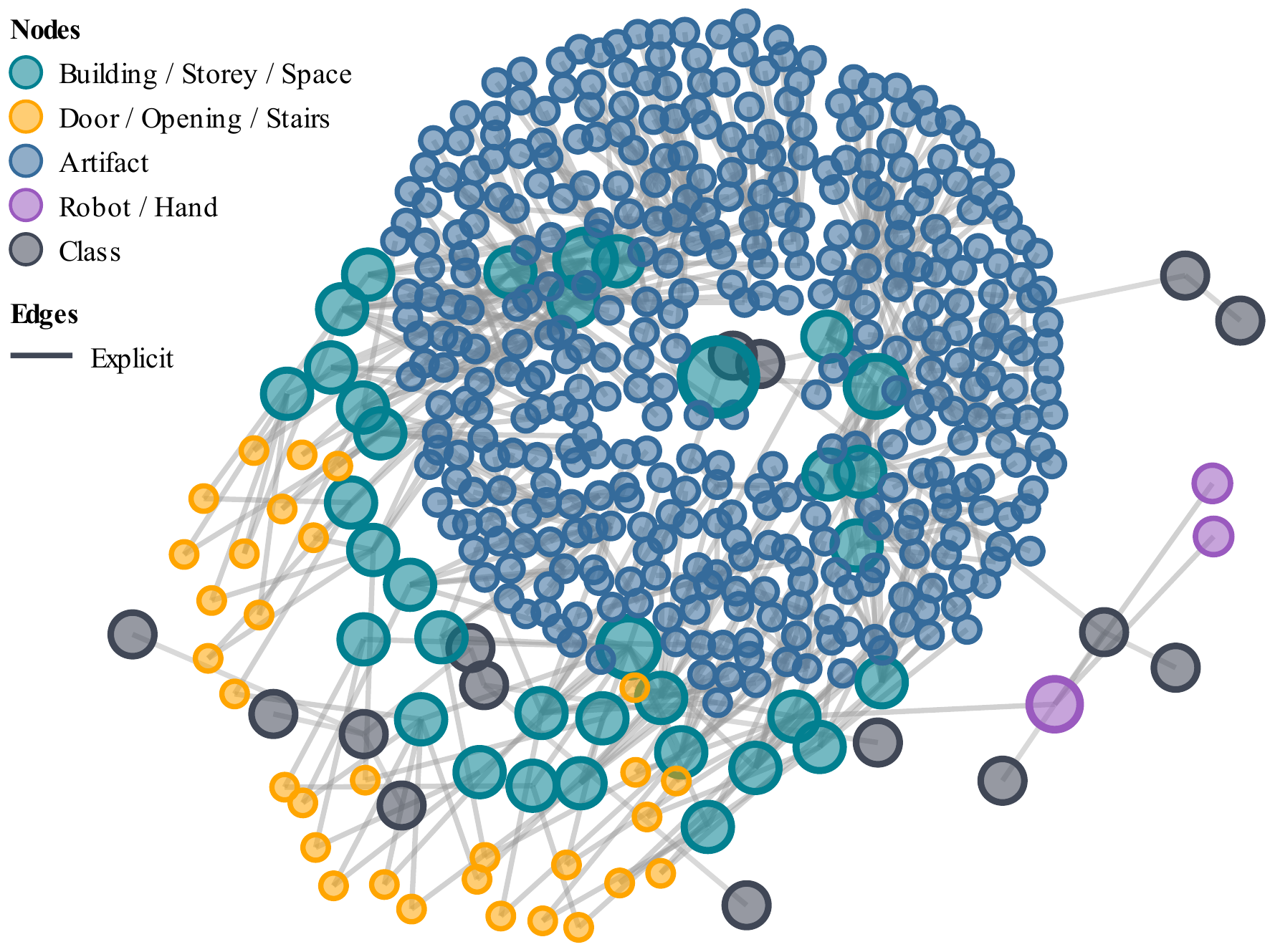}
\caption{Explicit (2,908 triples).}
\label{fig:wm-viz-explicit}
\end{subfigure}\hfill
\begin{subfigure}[t]{0.49\textwidth}
\centering
\includegraphics[width=0.714\linewidth]{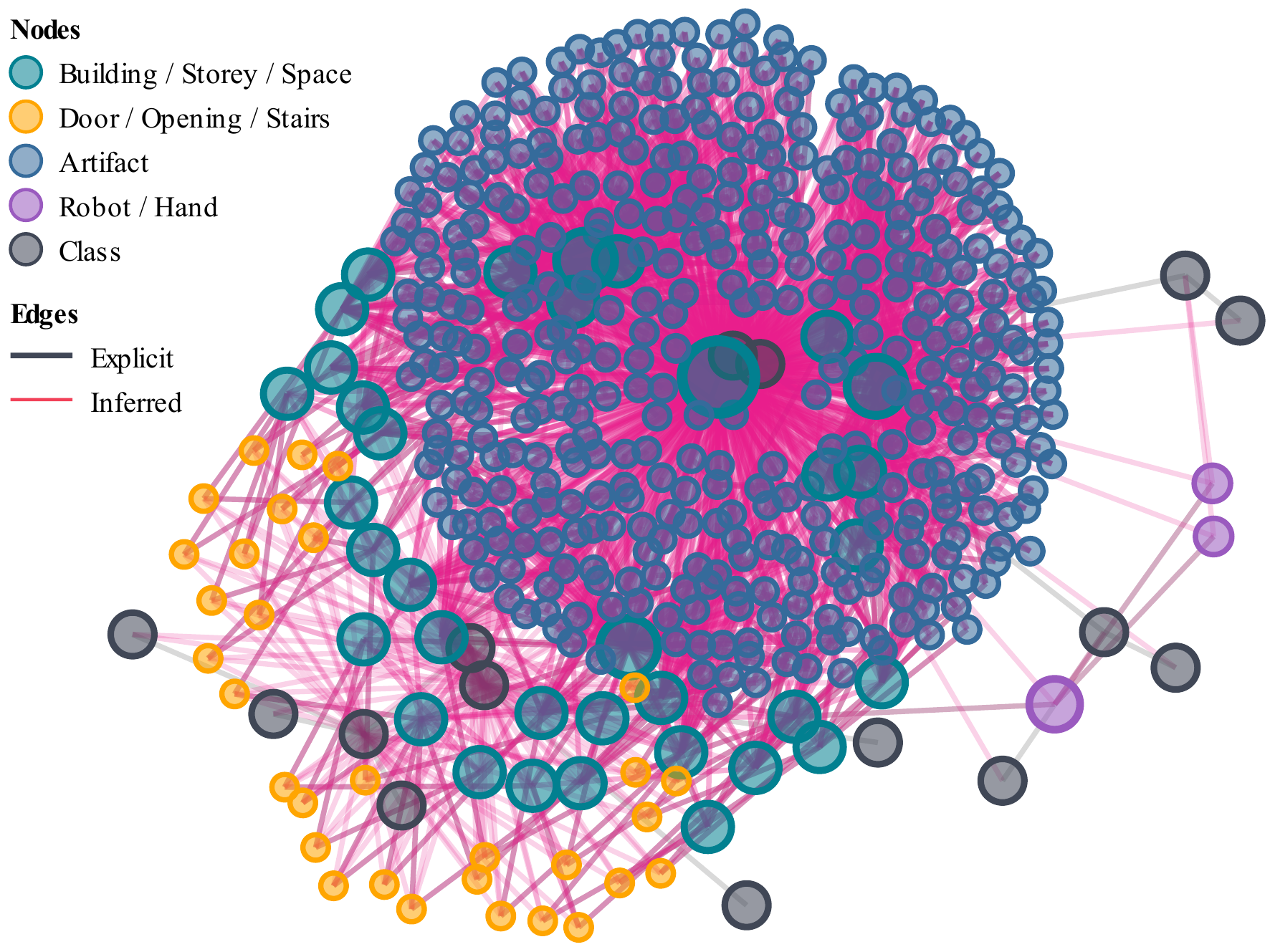}
\caption{Reasoned (19,394 triples; 16,486 inferred).}
\label{fig:wm-viz-reasoned}
\end{subfigure}
\caption{\textbf{World Model Before and After OWL Reasoning.} The reasoned view includes inferred facts in addition to explicit ones, making the representation substantially denser.}
\label{fig:wm-viz-comparison}
\end{minipage}
\end{center}

\newcommand{\wmadd}[1]{\texttt{+ #1}\par}
\newcommand{\wmdel}[1]{\texttt{- #1}\par}

\begin{center}
\begin{minipage}{\textwidth}
\centering
\scriptsize
\setlength{\tabcolsep}{2pt}
\captionsetup{type=table,skip=3pt}
\caption{\textbf{Representative World Model Updates and Inferred Facts.} For each action or symbolic observation event, the table lists directly edited dynamic-layer triples, selected inferred facts after OWL reasoning, and the axiom patterns responsible for those inferences. $+$ and $-$ denote inserted and deleted triples.}
\label{tab:plan-diffs}
{\renewcommand{\arraystretch}{0.90}
\begin{tabularx}{\textwidth}{@{}>{\raggedright\arraybackslash}p{0.12\textwidth}>{\raggedright\arraybackslash}X>{\raggedright\arraybackslash}X>{\raggedright\arraybackslash}p{0.17\textwidth}@{}}
\toprule
Event & Edited Triples & Inferred Facts & Axiom Basis \\
\midrule
\multicolumn{4}{@{}l}{\textit{Action Effects}} \\
\textsc{pick-one-hand}
&
\wmadd{holds(right\_hand, toothbrush\_213)}
\wmdel{isInsideOf(toothbrush\_213, storage\_cabinet\_55)}
&
\wmadd{carries(robot, toothbrush\_213)}
\wmadd{isCarriedBy(toothbrush\_213, robot)}
\wmadd{isHeldBy(toothbrush\_213, right\_hand)}
\wmdel{contains(storage\_cabinet\_55, toothbrush\_213)}
&
Inverse + carry/hold propagation
\\[2pt]
\textsc{move}\\
(carrying)
&
\wmadd{robotIsInSpace(robot, exercise\_room\_20)}
\wmdel{robotIsInSpace(robot, door\_1)}
&
\wmadd{isInSpace(robot, exercise\_room\_20)}
\wmadd{artifactIsInSpace(toothbrush\_213, exercise\_room\_20)}
\wmadd{isInStorey(toothbrush\_213, Floor\_A)}
\wmadd{isInBuilding(toothbrush\_213, Marstons)}
\wmdel{artifactIsInSpace(toothbrush\_213, door\_1)}
\wmdel{isInSpace(toothbrush\_213, door\_1)}
&
Spatial propagation + hierarchy
\\[2pt]
\textsc{place-in}
&
\wmadd{isInsideOf(toothbrush\_213, bucket\_36)}
\wmdel{holds(right\_hand, toothbrush\_213)}
&
\wmadd{contains(bucket\_36, toothbrush\_213)}
\wmdel{carries(robot, toothbrush\_213)}
\wmdel{isCarriedBy(toothbrush\_213, robot)}
\wmdel{isHeldBy(toothbrush\_213, right\_hand)}
&
Containment inverse + release cleanup
\\
\midrule
\multicolumn{4}{@{}l}{\textit{Symbolic Observation Update}} \\
\textsc{new object}
&
\wmadd{type(key\_new, Artifact)}
\wmadd{isInsideOf(key\_new, flower\_vase\_196)}
&
\wmadd{contains(flower\_vase\_196, key\_new)}
\wmadd{isInSpace(key\_new, bedroom\_8)}
\wmadd{artifactIsInSpace(key\_new, bedroom\_8)}
\wmadd{isInStorey(key\_new, Floor\_C)}
\wmadd{isInBuilding(key\_new, Marstons)}
\wmadd{hasArtifact(bedroom\_8, key\_new)}
\wmadd{spaceHasArtifact(bedroom\_8, key\_new)}
\wmadd{hasArtifact(Floor\_C, key\_new)}
\wmadd{hasArtifact(Marstons, key\_new)}
&
Containment inverse + hierarchy closure
\\
\bottomrule
\end{tabularx}
}
\end{minipage}
\end{center}

\newpage
\section{Agent System Prompts}
\label{app:agent-prompts}
\paragraph{Prompt notation.}
The prompts below are excerpted verbatim from the released code, with Markdown emphasis removed. Tokens formatted as \memorytoken{dialogue\_history} denote state fields injected at runtime, and \omittedtoken\ denotes omitted prompt text. In Figure~\ref{fig:system-architecture}, \memorytoken{tool\_history} groups \memorytoken{scene\_query\_history} and \memorytoken{pddl\_plan\_history}, and \memorytoken{task\_conditions} groups the \texttt{at\_end}, \texttt{always}, and \texttt{sometime} conditions.

\subsection{Flow Orchestrator Prompt}
\begin{promptbox}
{\footnotesize
\begin{alltt}
You are the Flow Orchestrator, the central orchestrator of the robot task
planning system.

## CONTEXT
\memorytoken{dialogue\_history}
\memorytoken{discovered\_objects}
\memorytoken{agent\_reports}

## YOUR ROLE
You manage the overall workflow by:
1. Determining user intent: Is this a question or a task?
2. Routing to appropriate agents: Decide which agent to invoke next
3. Providing intelligent responses: Answer questions with context and
   suggest follow-up tasks

## ROUTING
You can route to THREE agents:

### 1. Route to USER (Ask for clarification) \omittedtoken
### 2. Route to Scene Explorer (Explore environment) \omittedtoken
### 3. Route to Task Formalizer (Define constraints & goals) \omittedtoken

## DECISION LOGIC
Step 1: Classify user intent \omittedtoken
Step 2: Understand with common sense \omittedtoken
Step 3: Check if information is sufficient \omittedtoken
Step 4: Review system context \omittedtoken
Step 5: Make routing decision \omittedtoken

## OUTPUT
Always provide:
1. user_intent: "task_planning" or "question_answering"
2. next_agent: "USER" or "scene_explorer" or "task_formalizer"
3. reason: Clear explanation of why you chose this agent
4. question: (ONLY if routing to USER) The question or answer to provide
   - For question_answering: Include answer + optional task suggestion
   - For task_planning: Include clarification question if needed

Analyze the situation and decide the next agent to invoke.
\end{alltt}
}
\end{promptbox}

\newpage
\subsection{Scene Explorer Prompt}
\begin{promptbox}
\footnotesize
\begin{alltt}
You are the Scene Explorer. Think like someone who is physically in this
space trying to help -- use judgment and common sense, not just keyword
matching. Your goal is to give the planner a coherent, complete picture
of the scene.

## CONTEXT
\memorytoken{dialogue\_history}
\memorytoken{routing\_reason}
\memorytoken{space\_overview}
\memorytoken{scene\_query\_history}

## TOOLS

### semantic_search(query, reason, space_id, floor_id, object_id)
Natural language search with optional spatial filters.
- reason: one sentence explaining why you are making this call
  (required -- always fill this in).
- space_id: narrow to a specific room (e.g., "kitchen_1"). Do NOT use
  for room/space types -- rooms are not inside other rooms.
- floor_id: narrow to a floor (e.g., "Floor_B"). Also used with
  isConnectedTo to resolve spatial adjacency ("room next to X"). \omittedtoken

The hint field describes what the tool found and why -- \omittedtoken

### find_path(from_id, to_id, reason)
Returns shortest navigation path between two locations. Use when the task
specifies routing constraints (e.g., "nearest kitchen", "avoid corridor").
reason is required.

## STRATEGY
Be thorough. The task implies a full set of objects -- sources,
destinations, containers, portals, and spatial constraints. \omittedtoken

Call tools in parallel whenever possible. Start broad, then narrow:
1. Turn 1: search for all task-relevant objects simultaneously
2. Follow the hint if similar categories are needed or initial results
   are empty
3. Use object_id to find items inside containers once the container is
   identified
4. Call finish_exploration once all objects are found \omittedtoken

## EXAMPLES
Example 1: Object placement + container contents \omittedtoken
Example 2: Spatial adjacency \omittedtoken
Example 3: Portal search \omittedtoken

## OUTPUT
finish_exploration(exploration_summary="
Task: <one-line restatement>
Found:
  - <role>: `obj_id` (category) in `space_id`
  - <role>: `door_id` connects `space_a` <-> `space_b`
  ...
Not found: <anything not located>
Notes: <ambiguity resolutions only>
") \omittedtoken
\end{alltt}
\end{promptbox}

\newpage
\subsection{Task Formalizer Prompt}
\begin{promptbox}
\footnotesize
\begin{alltt}
You are the Task Formalizer, responsible for converting user tasks into
PDDL conditions.

## CONTEXT
\memorytoken{dialogue\_history}
\memorytoken{routing\_reason}
\memorytoken{discovered\_objects}

## YOUR ROLE
Convert user task instructions into 3 types of conditions based on PDDL
temporal planning:
1. at_end_condition (REQUIRED): What must be true when the task is complete
2. always_conditions (OPTIONAL): States that must be true throughout the
   entire execution
3. sometime_conditions (OPTIONAL): States that must be achieved at some
   point during execution (including high-level concepts like "make
   coffee", "bake bread")

## AVAILABLE PDDL PREDICATES
- (robotIsInSpace robot <space>) - Robot is in a room/space
- (isOnFloorOf <artifact> <space>) - Artifact on floor of a room/space
- (isInsideOf <artifact> <artifact>) - Artifact inside another artifact \omittedtoken
- (isOntopOf <artifact> <artifact>) - Artifact on top of another artifact \omittedtoken

## TOOLS
### 1. define_at_end_condition(goal_formula, reasoning) \omittedtoken
### 2. define_always_condition(predicate_formula, description) \omittedtoken
### 3. define_sometime_condition(predicate_formula, description) \omittedtoken

## WORKFLOW
1. Check objects: Do you have all necessary objects? If not ->
     request_more_exploration(reason)
2. Analyze task: What is the user really asking for?
3. Define end goal: What's the final state? -> define_at_end_condition (REQUIRED)
4. Identify constraints: Are there strict constraints that must hold
     during execution? -> define_always_condition
5. Identify intermediate states: Are there intermediate states that must
     occur during execution? -> define_sometime_condition

## CRITICAL RULES
1. Use EXACT object IDs from Available Objects - do NOT invent IDs. \omittedtoken
2. define_at_end_condition is MANDATORY -- Your response is INCOMPLETE
   without it. \omittedtoken

## OUTPUT
Tool calls in a single response:
  define_at_end_condition(goal_formula, reasoning)            [exactly 1, required]
  define_always_condition(predicate_formula, description)     [0 or more]
  define_sometime_condition(predicate_formula, description)   [0 or more]

Analyze the task and define all necessary conditions now.
\end{alltt}
\end{promptbox}

\newpage
\subsection{Planning Manager Prompt}
\begin{promptbox}
\footnotesize
\begin{alltt}
You are the Planning Manager - you arrange conditions from Task Formalizer
into ordered subgoals for PDDL planning.

## CONTEXT
\memorytoken{dialogue\_history}
\memorytoken{at\_end\_condition}
\memorytoken{always\_conditions}
\memorytoken{sometime\_conditions}
\memorytoken{pddl\_plan\_history}

## YOUR ROLE
CRITICAL: You are an ARRANGER, not a planner. Your job is to:
1. Take the conditions that Task Formalizer already defined
2. Arrange them in the correct execution order as subgoals
3. DO NOT add new intermediate steps or goals beyond what Task Formalizer
   specified \omittedtoken

## TOOLS
### 1. pddl_plan(subgoals, task_description) \omittedtoken
### 2. approve_plan(reasoning) \omittedtoken

## SUBGOAL CREATION
### Step 1: Convert Conditions to Subgoals
- Sometime Conditions -> Each becomes a subgoal \omittedtoken
- At-End Condition -> Handle carefully (see Step 2)
- Always Conditions -> Never convert to subgoals, only pass as
  always_conditions parameter
### Step 2: Handle At-End Condition
Default: Keep at_end as ONE subgoal (don't decompose it) \omittedtoken
### Step 3: Order the Subgoals \omittedtoken
### Step 4: DO NOT Add Extra Steps \omittedtoken

## SUBGOAL FORMAT
\{
  "goal_state": "(predicate formula)",  # Use ONLY predicates from list above
  "description": "What this achieves",
  "name": "lowercase_name",
  "order": 0  # Execution order
\}

## CRITICAL RULES
1. You are an ARRANGER, not a planner - Don't add intermediate steps
   beyond what Task Formalizer defined \omittedtoken
7. On retry: NEVER drop conditions - Every sometime condition and the
   at_end condition must always be present as subgoals. Only reorder or
   adjust formulas.

## OUTPUT
ReAct pattern -- Call ONE tool per response:
- First call: pddl_plan(subgoals=[...], task_description="...") \omittedtoken
- After pddl_plan result: \omittedtoken
  - Plan correctly achieves user intent -> approve_plan(reasoning="...")
  - Any subgoal FAILED, or plan is wrong/inappropriate for the task ->
    pddl_plan(...) again with corrected subgoals \omittedtoken

Now arrange the conditions from Task Formalizer into ordered subgoals.
\end{alltt}
\end{promptbox}

\newpage

\section{Dataset Scale Augmentation Visualization}
\label{app:dataset-scale-augmentation}

\begin{center}
\centering
\captionsetup{type=figure,skip=2pt}
\captionsetup[subfigure]{font=small,skip=1pt}
\begin{subfigure}[t]{\linewidth}
\centering
\includegraphics[width=\linewidth]{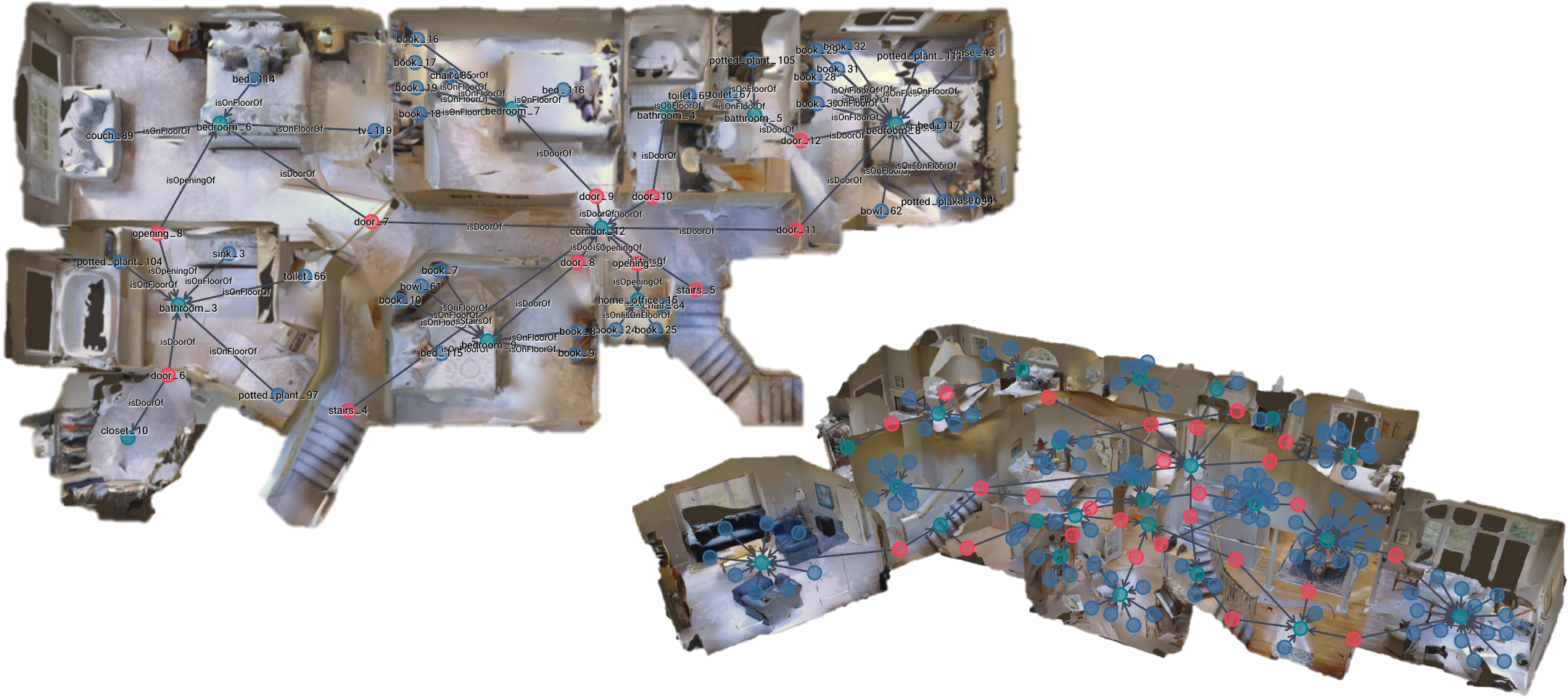}
\caption{Small. Original source scene graph (121 objects).}
\end{subfigure}
\vspace{0.2em}

\begin{subfigure}[t]{\linewidth}
\centering
\includegraphics[width=\linewidth]{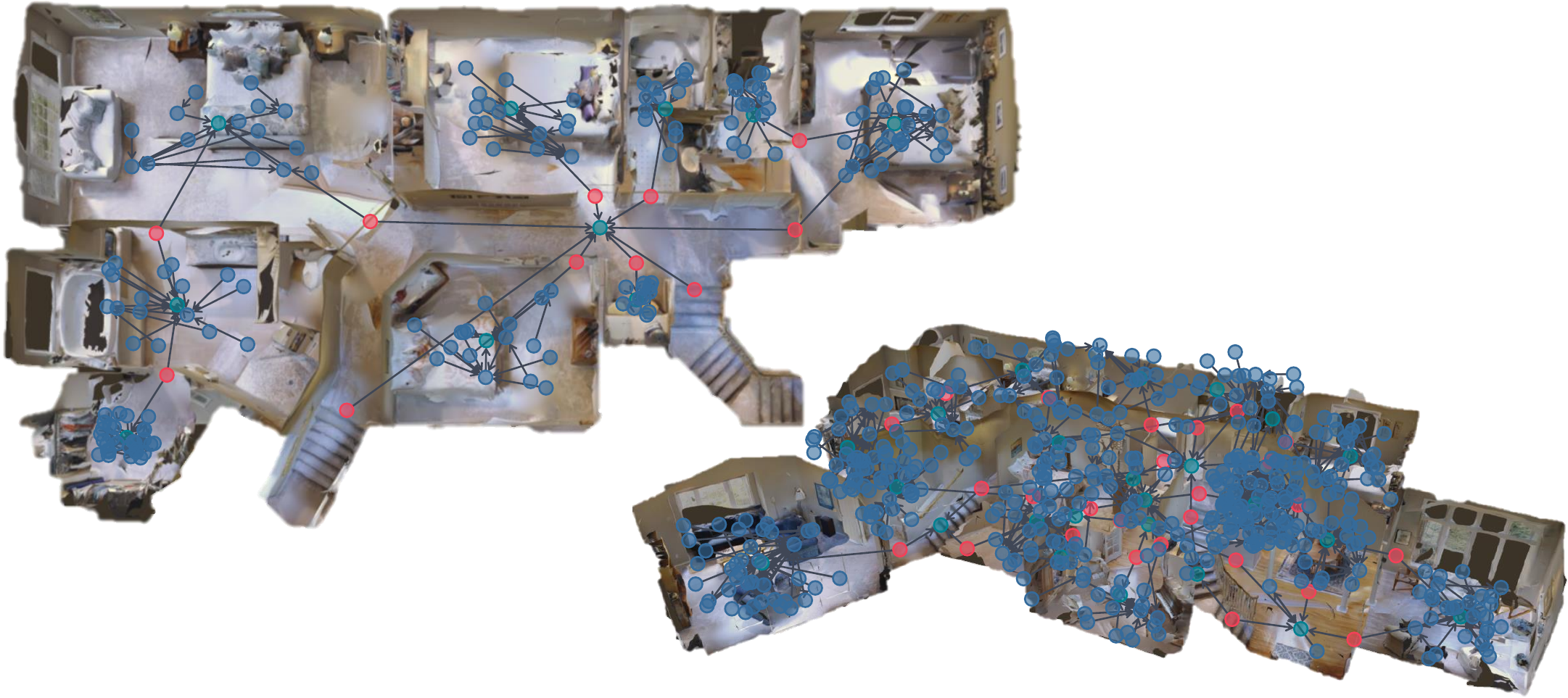}
\caption{Medium. Small with manually added objects (423 objects).}
\end{subfigure}
\vspace{0.2em}

\begin{subfigure}[t]{\linewidth}
\centering
\includegraphics[width=\linewidth]{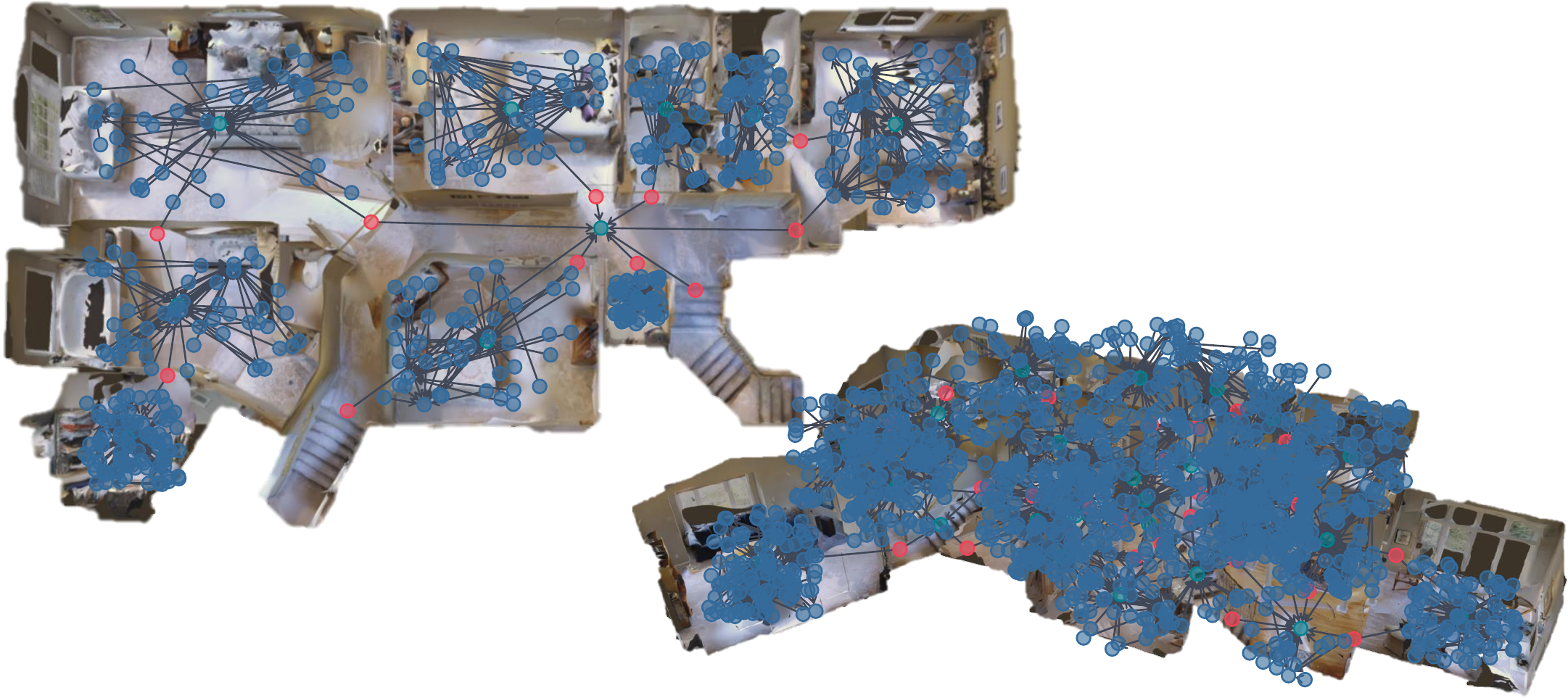}
\caption{Large. Medium with random augmentation to 3$\times$ objects (1269 objects).}
\end{subfigure}
\caption{\textbf{Dataset Scale Augmentation in Muleshoe.} Visualization of the same environment at the three scene scales. Each panel pairs a single-floor view on the left with a full-space view on the right.}
\label{fig:dataset-scale-augmentation}
\end{center}

\newpage

\section{General Instruction Examples}
\label{app:general-instruction-examples}
\enlargethispage{2\baselineskip}

\begin{center}
\centering
\scriptsize
\setlength{\tabcolsep}{3pt}
\captionsetup{type=table}
\caption{\textbf{General Instruction Examples for Klickitat.} Thirty instructions from the Small, Medium, and Large splits. \texttt{S}/\texttt{M}/\texttt{L} denote the splits, and gray arrow text shows derived evaluation conditions.}
\label{tab:appendix-general-instructions}
\begin{tabularx}{\textwidth}{cX}
\toprule
ID & Instruction \\
\midrule
\texttt{S01} & I think I left my cell phone in the dining room on Floor B. Can you find it and bring it to the home office on Floor C?
\annline{$\rightarrow$ (at\_end) \texttt{cell\_phone\_61} is in \texttt{home\_office\_18}.} \\
\texttt{S02} & There should be a bottle in the kitchen on Floor B. Take it out and put it on the couch in the living room on Floor B.
\annline{$\rightarrow$ (at\_end) \texttt{bottle\_3} is on \texttt{couch\_31} or \texttt{couch\_33}.} \\
\texttt{S03} & Move the sports ball from the playroom on Floor C to the home office on Floor C. Never go through the door connecting the lobby and the corridor on Floor B because it is under repair.
\annline{$\rightarrow$ (always) the robot does not pass through \texttt{door\_5}. (at\_end) \texttt{sports\_ball\_2} is in \texttt{home\_office\_18}.} \\
\texttt{S04} & Move the vase in the living room on Floor B to the bathroom on Floor C. Use only your right hand for this task.
\annline{$\rightarrow$ (always) \texttt{left\_hand} remains empty. (at\_end) \texttt{vase\_80} or \texttt{vase\_84} is in \texttt{bathroom\_3}, \texttt{bathroom\_4}, or \texttt{bathroom\_5}.} \\
\texttt{S05} & There is a book in one of the bedrooms on Floor C. Bring it to the living room on Floor B.
\annline{$\rightarrow$ (at\_end) \texttt{book\_78} is in \texttt{living\_room\_22} or \texttt{living\_room\_23}.} \\
\texttt{S06} & There aren't enough chairs in the dining room connected to the kitchen on Floor B. Can you bring one more chair from the other dining room on the same floor?
\annline{$\rightarrow$ (at\_end) at least one of \texttt{chair\_24}, \texttt{chair\_25}, or \texttt{chair\_26} is in \texttt{dining\_room\_16}.} \\
\texttt{S07} & Move the potted plant from the staircase on Floor B to the lobby. I'm afraid people might bump into it.
\annline{$\rightarrow$ (at\_end) \texttt{potted\_plant\_45} is in \texttt{lobby\_24}.} \\
\texttt{S08} & Move the vase from the bathroom on Floor C to the playroom. Use only your right hand.
\annline{$\rightarrow$ (always) \texttt{left\_hand} remains empty. (at\_end) \texttt{vase\_81} is in \texttt{playroom\_25}.} \\
\texttt{S09} & I need to move a chair from the dining room on Floor B to the bedroom on Floor A. Please use the stairs leading to Floor A.
\annline{$\rightarrow$ (sometime) robot passes through \texttt{stairs\_7}. (at\_end) at least one chair from \texttt{chair\_18}--\texttt{chair\_22} or \texttt{chair\_24}--\texttt{chair\_26} is in \texttt{bedroom\_6}.} \\
\texttt{S10} & Move one potted plant from the bedroom connected to the closet on Floor C to another bedroom. And make sure to close the door when you leave the room.
\annline{$\rightarrow$ (at\_end) at least one of \texttt{potted\_plant\_46} or \texttt{potted\_plant\_47} is in \texttt{bedroom\_6}, \texttt{bedroom\_8}, or \texttt{bedroom\_9}, and \texttt{door\_6} is closed.} \\
\midrule
\texttt{M01} & Please turn on the coffee machine in the kitchen on Floor B. Also, bring the mug from the tray in the bedroom on Floor C and place it on the coffee machine.
\annline{$\rightarrow$ (at\_end) \texttt{coffee\_machine\_170} is switched on, and \texttt{mug\_71} is on it.} \\
\texttt{M02} & Turn on the TV in the living room on Floor B and gather the cushions scattered around the room onto the sofa.
\annline{$\rightarrow$ (at\_end) \texttt{tv\_211} is switched on, and cushions \texttt{cushion\_219}--\texttt{cushion\_223} are on \texttt{sofa\_208}.} \\
\texttt{M03} & I think I left the refrigerator in the kitchen open. Could you take the juice out of it and place it on the dining table? And please close the refrigerator when you are done.
\annline{$\rightarrow$ (at\_end) \texttt{juice\_357} is on \texttt{dining\_table\_112} or \texttt{dining\_table\_131}, and \texttt{refrigerator\_164} is closed.} \\
\texttt{M04} & Turn on the laptop in the home office on Floor C and take the hole punch out of the storage cabinet onto the desk.
\annline{$\rightarrow$ (at\_end) \texttt{laptop\_144} is switched on, and \texttt{hole\_punch\_323} is on \texttt{desk\_139}.} \\
\texttt{M05} & Take the jeans out of the dresser in the closet on Floor C and put them on the armchair in the bedroom.
\annline{$\rightarrow$ (at\_end) \texttt{jeans\_294} is on \texttt{armchair\_48} or \texttt{armchair\_49}.} \\
\texttt{M06} & Take the wine glasses out of the cupboard in the dining room on Floor B and set them on the dining table.
\annline{$\rightarrow$ (at\_end) \texttt{wine\_glass\_121} and \texttt{wine\_glass\_122} are on \texttt{dining\_table\_112}.} \\
\texttt{M07} & Turn on the microwave in the kitchen on Floor B. And if the door leading to the dining room is closed, please open it.
\annline{$\rightarrow$ (at\_end) \texttt{microwave\_169} is switched on, and \texttt{door\_3} is open.} \\
\texttt{M08} & Cook the chicken from the refrigerator in the kitchen on Floor B in the oven, then serve it on the island table in the kitchen.
\annline{$\rightarrow$ (sometime) \texttt{chicken\_353} is inside \texttt{oven\_161} while it is closed and switched on. (at\_end) it is on \texttt{island\_table\_158}.} \\
\texttt{M09} & Heat up the beef from the refrigerator in the kitchen on Floor B in the microwave, then place it on the island table in the kitchen.
\annline{$\rightarrow$ (sometime) \texttt{beef\_354} is inside \texttt{microwave\_169} while it is closed and switched on. (at\_end) it is on \texttt{island\_table\_158}.} \\
\texttt{M10} & Wash the bowl from the storage cabinet in the kitchen on Floor A in the dishwasher, then put it back on top of the storage cabinet.
\annline{$\rightarrow$ (sometime) \texttt{bowl\_341} is inside \texttt{dishwasher\_156} while it is closed and switched on. (at\_end) it is on \texttt{storage\_cabinet\_154}.} \\
\midrule
\texttt{L01} & Turn on the microwave in the kitchen on Floor B. But since the corridor is being cleaned, please go through the living room.
\annline{$\rightarrow$ (always) the robot does not pass through \texttt{corridor\_12}. (at\_end) \texttt{microwave\_169} is switched on.} \\
\texttt{L02} & We need one more chair in the dining room on Floor A. Can you bring a chair from the dining room next to the corridor on Floor B?
\annline{$\rightarrow$ (at\_end) \texttt{chair\_132} or \texttt{chair\_133} is in \texttt{dining\_room\_15}.} \\
\texttt{L03} & Turn on the TV in the living room on Floor B. But please avoid the stairs connecting the living room and the lobby on Floor B.
\annline{$\rightarrow$ (always) the robot does not pass through \texttt{stairs\_4}. (at\_end) \texttt{tv\_211} is switched on.} \\
\texttt{L04} & There should be a hole punch in the storage cabinet in the home office on Floor C. Take it out and put it on the desk.
\annline{$\rightarrow$ (at\_end) \texttt{hole\_punch\_323} is on \texttt{desk\_139}.} \\
\texttt{L05} & The lamps in the bedroom with a bathroom right next to you are on. Could you turn them all off so I can sleep comfortably?
\annline{$\rightarrow$ (at\_end) \texttt{lamp\_83} and \texttt{lamp\_84} are switched off.} \\
\texttt{L06} & Take the milk from the refrigerator in the kitchen on Floor B and put it on the dining room table.
\annline{$\rightarrow$ (at\_end) \texttt{milk\_348} is on \texttt{dining\_table\_112} or \texttt{dining\_table\_131}.} \\
\texttt{L07} & Take the jeans out of the closet dresser on Floor C and lay them on the bed.
\annline{$\rightarrow$ (at\_end) \texttt{jeans\_294} is on \texttt{bed\_53} or \texttt{bed\_73}.} \\
\texttt{L08} & Cook the cheese from the refrigerator in the kitchen on Floor B in the oven, then serve it on the island table in the kitchen.
\annline{$\rightarrow$ (sometime) \texttt{cheese\_350} is inside \texttt{oven\_161} while it is closed and switched on. (at\_end) it is on \texttt{island\_table\_158}.} \\
\texttt{L09} & Heat up the milk from the refrigerator in the kitchen on Floor B in the microwave, then place it on the island table in the kitchen.
\annline{$\rightarrow$ (sometime) \texttt{milk\_348} is inside \texttt{microwave\_169} while it is closed and switched on. (at\_end) it is on \texttt{island\_table\_158}.} \\
\texttt{L10} & Wash the cup from the storage cabinet in the kitchen on Floor A in the dishwasher, then put it back on top of the storage cabinet.
\annline{$\rightarrow$ (sometime) \texttt{cup\_343} is inside \texttt{dishwasher\_156} while it is closed and switched on. (at\_end) it is on \texttt{storage\_cabinet\_154}.} \\
\bottomrule
\end{tabularx}
\end{center}

\newpage

\section{Experimental Protocol Details}
\label{app:experimental-protocol-details}

\subsection{Controlled Implementation Settings}
This section summarizes the implementation settings used in the main comparison. Across methods, we use the same task instances, scene facts, initial symbolic states, LLM setting, primitive action set, and symbolic evaluation procedure. Differences are kept only where they are part of the original method design.

\paragraph{LLM setting.}
All methods use GPT-4o with temperature 0; \textsc{OntoPlan} and its ablations use the \texttt{gpt-4o-2024-08-06} snapshot. In addition, the \texttt{semantic\_search} function of \textsc{OntoPlan} embeds object categories with OpenAI \texttt{text-embedding-3-small} (256 dimensions) and retrieves them with an HNSW index. The world model is served by GraphDB 10.7 with the OWL 2 RL ruleset. All LLM calls are made through the OpenAI API, and GraphDB and Fast Downward run locally on a single CPU workstation.

\paragraph{Scene representation.}
All methods start from the same scene facts and initial symbolic state. The scene facts include spaces, connectivity, object instances, object locations, object states, and the initial robot state. SayPlan and DELTA receive these facts as serialized text. \textsc{OntoPlan} loads the same facts into the ontology-grounded world model. Thus, the comparison varies the scene interface, not the underlying scene information. All scene facts are instantiated before any LLM call through deterministic mappings from the benchmark scene descriptions. None of the evaluated methods receives raw sensor input, and no LLM or learned perception model is used to construct the initial symbolic state.

\paragraph{Primitive action set.}
All methods use the same 16 primitive actions for task planning and evaluation: move, access, open/close for items and doors, power-on/power-off, and one-/two-handed variants of pick, place-to-location, place-in, and place-on. SayPlan is prompted to generate plans using this primitive action set. DELTA and \textsc{OntoPlan} use the same action set through the shared PDDL domain. All returned plans are evaluated with the same action names, preconditions, effects, and costs.

\paragraph{PDDL planning setup.}
\textsc{OntoPlan} uses the shared PDDL domain directly through the \textit{PDDL plan} tool. Following its original design, DELTA receives the same domain as reference context and generates its own planner-facing domain and problem files. Both methods use Fast Downward with \texttt{lazy\_wastar}, the FF heuristic, and search weight 2.

\paragraph{Exploration and planning calls.}
SayPlan and \textsc{OntoPlan} are each allowed up to 20 scene-exploration LLM calls and 5 plan-generation calls. DELTA follows its fixed workflow and does not use repeated exploration or plan-generation calls.

\subsection{Symbolic Evaluation Procedure}
All returned action sequences are evaluated with the same symbolic evaluator.
Starting from the task initial symbolic state, the evaluator applies actions sequentially, checks each action's preconditions, and updates the symbolic state using the shared action effects.
Execution stops after the final action or at the first precondition failure.
\texttt{always} conditions are checked in the initial state and after every action, \texttt{sometime} and \texttt{ordering} conditions are tracked over the executed actions, and \texttt{at\_end} is checked in the reached state.
The reached state and execution trajectory are then used to compute the metrics defined in Section~\ref{sec:evaluation-metrics}.

\clearpage
\section{Statistical Analysis}
\label{app:statistical-significance}

This section reports paired bootstrap 95\% confidence intervals for the general-task comparisons with the baselines and for the ablation comparisons. For the baselines, we report \textsc{OntoPlan}-minus-baseline differences in the four binary performance metrics from Table~\ref{tab:robot-task-planning}: \textit{Plan Returned}, \textit{Plan Executable}, \textit{Goal Reached}, and \textit{Task Success}. All intervals are computed using bootstrap resampling stratified by scene scale.

\subsection{Analysis Setup}
Each bootstrap resample draws 50 tasks with replacement from each scene scale, preserving the benchmark's 1:1:1 Small/Medium/Large composition. We use 10,000 bootstrap resamples. Within each resample, the same task instances are used for \textsc{OntoPlan} and each baseline, yielding paired bootstrap 95\% confidence intervals. For each metric, we report the rate difference (\textsc{OntoPlan} minus baseline). Confidence intervals are computed from the 2.5th and 97.5th percentiles of the bootstrap distribution. This analysis quantifies task-sampling uncertainty while keeping the scene-scale composition fixed.

\subsection{Paired Comparisons}
Table~\ref{tab:statistical-significance-overall} reports these intervals for \textsc{OntoPlan} versus each baseline across the four performance metrics.

\begin{center}
\begin{minipage}{\textwidth}
\centering
\small
\setlength{\tabcolsep}{3pt}
\captionsetup{type=table,skip=3pt}
\caption{\textbf{Paired bootstrap 95\% confidence intervals for general-task performance.} Intervals are computed using bootstrap resampling stratified by scene scale. Differences are computed as \textsc{OntoPlan} minus baseline across the 150 matched general-task instances.}
\label{tab:statistical-significance-overall}
\begin{tabularx}{\textwidth}{@{}>{\raggedright\arraybackslash}p{0.12\textwidth}>{\raggedright\arraybackslash}p{0.24\textwidth}>{\raggedright\arraybackslash}X>{\raggedright\arraybackslash}X@{}}
\toprule
Baseline & Metric & \shortstack[l]{Difference\\(\textsc{OntoPlan} $-$ baseline)} & 95\% CI \\
\midrule
\multirow{4}{*}{SayPlan} & Plan Returned & 0.33 & [0.25, 0.41] \\
& Plan Executable & 0.61 & [0.53, 0.68] \\
& Goal Reached & 0.58 & [0.49, 0.66] \\
& Task Success & 0.61 & [0.53, 0.69] \\
\midrule
\multirow{4}{*}{DELTA} & Plan Returned & 0.55 & [0.47, 0.63] \\
& Plan Executable & 0.83 & [0.77, 0.89] \\
& Goal Reached & 0.73 & [0.65, 0.80] \\
& Task Success & 0.75 & [0.68, 0.82] \\
\bottomrule
\end{tabularx}
\end{minipage}
\end{center}

\subsection{Ablation Comparisons}
Table~\ref{tab:statistical-significance-ablation} applies the same paired, scale-stratified bootstrap to the full system and the two ablation variants in Table~\ref{tab:tool-ablation}, together with exact McNemar tests on per-task Task Success.

\begin{center}
\begin{minipage}{\textwidth}
\centering
\small
\setlength{\tabcolsep}{3pt}
\captionsetup{type=table,skip=3pt}
\caption{\textbf{Paired comparisons between the full system and the ablation variants.} Differences are computed as full system minus variant across the 150 matched general-task instances. Discordant pairs count tasks solved only by the full system and only by the variant.}
\label{tab:statistical-significance-ablation}
\begin{tabularx}{\textwidth}{@{}>{\raggedright\arraybackslash}p{0.2\textwidth}YYYY@{}}
\toprule
Variant & \shortstack{Task Success\\difference} & 95\% CI & \shortstack{Discordant\\pairs} & \shortstack{McNemar\\$p$} \\
\midrule
w/o \textit{Scene query} & 0.02 & [$-$0.04, 0.09] & 13 / 10 & 0.68 \\
w/o \textit{PDDL plan} & 0.57 & [0.49, 0.65] & 87 / 1 & $<10^{-24}$ \\
\bottomrule
\end{tabularx}
\end{minipage}
\end{center}

\newpage
\section{Failure Analysis of \textsc{OntoPlan}}
\label{app:failure-analysis}
\textsc{OntoPlan} fails 17 of the 150 general tasks. Every returned plan executes without a precondition failure (one failed task returns no plan), and the \textit{PDDL plan} tool succeeds on its first call in 145 of the 150 tasks. Comparing the conditions written by the Task Formalizer with the benchmark conditions, Table~\ref{tab:failure-analysis} groups the failures by their origin.

\begin{center}
\begin{minipage}{\textwidth}
\centering
\small
\setlength{\tabcolsep}{3pt}
\captionsetup{type=table,skip=3pt}
\caption{\textbf{Failure categories of \textsc{OntoPlan} on the general tasks.} Tasks are given as environment and task ID.}
\label{tab:failure-analysis}
\begin{tabularx}{\textwidth}{@{}>{\raggedright\arraybackslash}p{0.36\textwidth}c>{\raggedright\arraybackslash}X@{}}
\toprule
Category & Count & Tasks \\
\midrule
Referring expression grounded to the wrong object or place & 13 & Klickitat M03, S10; Lindenwood L08, L09, M10, S08; Marstons L03, L04, L05, L07, M10, S10; Muleshoe L07 \\
Part of the goal omitted & 2 & Lakeville M07; Marstons S07 \\
Trajectory constraint omitted & 1 & Klickitat L01 \\
No valid goal produced & 1 & Muleshoe M06 \\
\bottomrule
\end{tabularx}
\end{minipage}
\end{center}

Grounding errors include choosing a different instance of the right category (e.g., \texttt{dining\_table\_58} instead of \texttt{dining\_table\_57} in Lindenwood S08) and resolving a description to the wrong space (e.g., ``the study'' to \texttt{bedroom\_7} instead of \texttt{home\_office\_21} in Marstons L04). Omissions include leaving out some of the doors to open in ``open all closed doors'' instructions and not formalizing the instruction to avoid the corridor in Klickitat L01. These errors arise in instruction interpretation and scene grounding rather than in planning, which motivates verification of grounded referents and formalized conditions as future work.

\section{Unreturned Plans Across Methods}
\label{app:unreturned-plans}
Table~\ref{tab:unreturned-plans} attributes every general task for which a method returned no plan to the stage at which it stopped. \textsc{OntoPlan} and DELTA obtain plans from Fast Downward with the same configuration and a 60\,s limit per subgoal, whereas SayPlan generates plans with the LLM after scene-graph search, so their failures arise at different stages. For DELTA, we take the first sub-problem it could not solve and check whether it parses, whether its predicates match the domain in name, arity, and argument types, and whether every object it mentions is declared. If the sub-problem is well formed but unsolvable, we add the true scene facts over its declared objects, and then the missing scene objects with their facts, and record which addition makes it solvable. For SayPlan, we identify the condition that ended its scene-graph search. No unreturned plan was caused by the planner time limit.

\begin{center}
\begin{minipage}{\textwidth}
\centering
\small
\setlength{\tabcolsep}{3pt}
\captionsetup{type=table,skip=3pt}
\caption{\textbf{Causes of unreturned plans on the 150 general tasks.} Counts are numbers of tasks.}
\label{tab:unreturned-plans}
\begin{tabularx}{\textwidth}{@{}>{\raggedright\arraybackslash}p{0.16\textwidth}>{\raggedright\arraybackslash}Xc@{}}
\toprule
Method & Cause & Count \\
\midrule
\textsc{OntoPlan} & No valid goal produced & 1 \\
\midrule
\multirow{4}{*}{DELTA} & Objects used in the sub-problem not declared & 61 \\
& Initial-state facts needed to reach the subgoal missing & 18 \\
& Predicate applied to arguments of the wrong type & 4 \\
& Free variable in the goal & 1 \\
\midrule
\multirow{3}{*}{SayPlan} & Invalid scene-graph search command & 44 \\
& Search budget of 20 calls exhausted & 4 \\
& Context window of the LLM exceeded & 3 \\
\bottomrule
\end{tabularx}
\end{minipage}
\end{center}

DELTA's failures stem from incomplete or malformed planner-facing specifications. In 48 of the 61 tasks with undeclared objects, the undeclared objects are rooms, doors, or stairs that the generated problem refers to without declaring; the misused predicates place an object inside a room or inside a surface without containment. SayPlan's failures occur during scene-graph search: in 30 of the 44 tasks with an invalid command, the LLM tries to expand or contract an object node instead of a room, and in the remaining tasks it issues an action or a malformed command. The three context-window failures occur in Large scenes, where the accumulated search prompt exceeds the 128k-token context window of GPT-4o.

\newpage

\section{Full Results on Ambiguous Instructions}
\label{app:ambiguous-full-results}

\begin{center}
\begin{minipage}{\textwidth}
\centering
\footnotesize
\setlength{\tabcolsep}{3pt}
\captionsetup{type=table}
\caption{\textbf{Per-Instruction Results on Ambiguous Instructions.} Outcomes for all 10 ambiguous instructions with the fixed automatic reply versus direct user replies to clarification questions. \cmark/\xmark\ denote whether the goal is reached, and gray arrow text summarizes the interpreted user intent.}
\label{tab:appendix-ambiguous-full}
\begin{tabularx}{\textwidth}{>{\centering\arraybackslash}p{0.03\textwidth}X>{\centering\arraybackslash}p{0.06\textwidth}@{}>{\centering\arraybackslash}p{0.06\textwidth}X}
\toprule
ID & Instruction & \multicolumn{2}{c}{Goal Reached} & Interaction Log \\
\cmidrule(lr){3-4}
 &  & Auto & User &  \\
\midrule
\texttt{A01} & Now that the guests have left, put the coffee cups in the living room back where they originally were.
\annline{$\rightarrow$ return \texttt{coffee\_cup\_228}, \texttt{coffee\_cup\_229}, and \texttt{coffee\_cup\_230} to \texttt{kitchen\_20}.} & \xmark & \cmark & \textbf{\textsc{OntoPlan}:} Where were the coffee cups originally placed in the living room? \newline \textbf{User:} Kitchen on Floor\_B. \\
\texttt{A02} & Bring me the new juice we bought yesterday so I can try it in the Floor A living room.
\annline{$\rightarrow$ move \texttt{juice\_357} to \texttt{living\_room\_21}.} & \cmark & \cmark & \textbf{\textsc{OntoPlan}:} Which juice is the new one you bought yesterday? Is it \texttt{juice\_357} (orange plastic) in \texttt{kitchen\_20} or \texttt{juice\_381} (green wooden) in \texttt{bathroom\_3}? \newline \textbf{User:} I mean \texttt{juice\_357}. \\
\texttt{A03} & Bring me the vase Anna gifted me to my bedroom so I can take a photo.
\annline{$\rightarrow$ move \texttt{vase\_373} to \texttt{bedroom\_8}.} & \xmark & \cmark & \textbf{\textsc{OntoPlan}:} Is \texttt{vase\_373} (located in \texttt{living\_room\_21}) the vase Anna gifted you? Also, which bedroom is yours: \texttt{bedroom\_6} (Floor\_A), \texttt{bedroom\_7}, \texttt{bedroom\_8}, or \texttt{bedroom\_9} (Floor\_C)? \newline \textbf{User:} Yes, it is. My bedroom is number 8. \\
\texttt{A04} & Bring the book I was reading yesterday from the living room to my room.
\annline{$\rightarrow$ move \texttt{book\_203} to \texttt{bedroom\_8}.} & \xmark & \cmark & \textbf{\textsc{OntoPlan}:} Which book were you reading yesterday? \texttt{book\_202} (white, \texttt{living\_room\_22}), \texttt{book\_203} (beige, \texttt{living\_room\_22}), or \texttt{book\_204} (beige, \texttt{living\_room\_22})? And which bedroom is your room? \texttt{bedroom\_6} (Floor\_A), \texttt{bedroom\_7}, \texttt{bedroom\_8}, or \texttt{bedroom\_9} (Floor\_C)? \newline \textbf{User:} The book is \texttt{book\_203}, and my room is \texttt{bedroom\_8}. \\
\texttt{A05} & We finished eating, so please clean up the desk in the dining room.
\annline{$\rightarrow$ move \texttt{sugar\_bowl\_130}, \texttt{pepper\_shaker\_129}, and \texttt{salt\_shaker\_128} into \texttt{cupboard\_119}.} & \xmark & \cmark & \textbf{\textsc{OntoPlan}:} I found \texttt{sugar\_bowl\_130}, \texttt{pepper\_shaker\_129}, and \texttt{salt\_shaker\_128} on \texttt{desk\_127} in the dining room. What should I do with each item after cleaning them up? \newline \textbf{User:} Move all of them to cupboard in same room. \\
\texttt{A06} & Bring the thing that broke while cooking yesterday to the lobby.
\annline{$\rightarrow$ move \texttt{oven\_161} to \texttt{lobby\_24}.} & \xmark & \cmark & \textbf{\textsc{OntoPlan}:} Which object broke while cooking yesterday? Is it the \texttt{kitchen\_drawer\_167} in \texttt{kitchen\_20} or the \texttt{vegetable\_basket\_355} in \texttt{kitchen\_20}? \newline \textbf{User:} It's oven. \\
\texttt{A07} & Put back the shoes used as a design sample in the home office to their original place.
\annline{$\rightarrow$ return \texttt{shoes\_149} to \texttt{bookshelf\_140}.} & \xmark & \cmark & \textbf{\textsc{OntoPlan}:} Which closet should I put the shoes back in? \texttt{closet\_10} or \texttt{closet\_11} on Floor\_C? \newline \textbf{User:} Return it to bookshelf. \\
\texttt{A08} & There are two pillows in John's room. He doesn't like one of them. Swap it with one of the pillows in my room.
\annline{$\rightarrow$ swap \texttt{pillow\_82} and \texttt{pillow\_66} between \texttt{bed\_53} and \texttt{bed\_73}.} & \xmark & \xmark & \textbf{\textsc{OntoPlan}:} There are no pillows found in John's room (\texttt{playroom\_25}). Which pillow from your room (\texttt{bedroom\_7}) would you like to swap? \newline \textbf{User:} John's room is not playroom. His room is bedroom. Please move the pillow. \\
\texttt{A09} & Plates on the floor are dangerous if they break, so put them back where they originally were.
\annline{$\rightarrow$ move \texttt{plate\_137} and \texttt{plate\_110} into \texttt{cupboard\_134}.} & \xmark & \xmark & \textbf{\textsc{OntoPlan}:} Where were the plates originally placed? \newline \textbf{User:} Put both into cupboard in same room. \\
\texttt{A10} & Please tidy up the desk in the living room.
\annline{$\rightarrow$ move \texttt{mug\_205}, \texttt{mug\_206}, \texttt{book\_203}, and \texttt{book\_204} into \texttt{wall\_cabinet\_191}.} & \xmark & \cmark & \textbf{\textsc{OntoPlan}:} I found \texttt{book\_203}, \texttt{book\_204}, \texttt{mug\_205}, and \texttt{mug\_206} on \texttt{desk\_193} in the living room. Where should I place each item after tidying? \newline \textbf{User:} Move \texttt{mug\_205}, \texttt{mug\_206}, \texttt{book\_203}, and \texttt{book\_204} into any wall cabinet. \\
\bottomrule
\end{tabularx}
\end{minipage}
\end{center}

\newpage

\section{Full Results on Infeasible Instructions}
\label{app:infeasible-full-results}

{%
\footnotesize
\setlength{\tabcolsep}{3pt}
\setlength{\LTleft}{0pt}
\setlength{\LTright}{0pt}
\begin{xltabular}{\textwidth}{@{}>{\centering\arraybackslash}p{0.03\textwidth}>{\raggedright\arraybackslash}X>{\centering\arraybackslash}p{0.15\textwidth}>{\raggedright\arraybackslash}X@{}}
\caption{\textbf{Per-Instruction Results on Infeasible Instructions.} Outcomes for all 10 infeasible instructions in the Klickitat Medium environment. Gray arrow text summarizes the infeasibility rationale.}
\label{tab:appendix-infeasible-full}\\
\toprule
ID & Instruction & Response Type & System Output \\
\midrule
\endfirsthead
\caption[]{\textbf{Per-Instruction Results on Infeasible Instructions.} (continued)}\\
\toprule
ID & Instruction & Response Type & System Output \\
\midrule
\endhead
\bottomrule
\endfoot
\bottomrule
\endlastfoot
\texttt{I01} & I want to skateboard again after a long time. Could you find the skateboard in the living room on the B floor and bring it to the lobby? \annline{$\rightarrow$ no skateboard in the environment.} & Clarification\newline Request & \textit{We couldn't find the skateboard in either of the living rooms on the B floor. Could you please confirm if it might be in another location or provide more details about where it might be?} \\
\texttt{I02} & I need an extra chair for the guests. Please pick up the chair in the kitchen on the A floor and bring it to the lobby. \annline{$\rightarrow$ no chair in the kitchen on Floor A.} & Clarification\newline Request & \textit{No chairs were found in the kitchen on Floor A. Could you please specify another location where I might find a chair?} \\
\texttt{I03} & I'm thirsty after doing laundry. Could you move to the laundry room and find the juice, then bring it to the kitchen on the B floor? \annline{$\rightarrow$ no laundry room in the environment.} & Clarification\newline Request & \textit{The juice is already located in the kitchen on Floor B. Do you need anything else, or is there another task you intended?} \\
\texttt{I04} & I want to watch TV in the bedroom. Please bring the remote control that is on the side table in the living room on the B floor to the bedroom on the C floor. \annline{$\rightarrow$ no remote control on the side table (none exists in the environment).} & Clarification\newline Request & \textit{We couldn't find the remote control on the side tables in the living rooms on the B floor. Could you please confirm its location or provide more details?} \\
\texttt{I05} & I have a business trip coming up. Could you take the briefcase out of the storage cabinet in the home office on the C floor and bring it to the lobby on the B floor? \annline{$\rightarrow$ no briefcase in the storage cabinet (none exists in the environment).} & Clarification\newline Request & \textit{Could you provide more details about the briefcase? For example, its color, size, or any distinguishing features that might help us locate it.} \\
\texttt{I06} & I'd like to read before bed. Please turn on the book in the bedroom on the C floor. \annline{$\rightarrow$ \textit{book} is not an operable device.} & Misdirected\newline Plan & \textbf{Plan interpretation:} Reinterpreted as a lamp-related request and generated a plan to turn on \texttt{lamp\_85}. \\
\texttt{I07} & The kitchen is too cramped. Could you put the refrigerator inside the microwave in the kitchen on the B floor? \annline{$\rightarrow$ physical containment impossible due to size constraints.} & Clarification\newline Request & \textit{It seems impractical to put the refrigerator inside the microwave due to size constraints. Could you clarify what you would like to achieve in the kitchen? Perhaps we can find another way to manage the space.} \\
\texttt{I08} & I need that book for the meeting tomorrow. Could you bring the book from the living room on the B floor to the bedroom on the C floor? The baby is on the stairs, so please don't use the stairs. \annline{$\rightarrow$ only feasible path requires stairs, which is disallowed.} & Clarification\newline Request & \textit{The only available path to move the book involves using the stairs, which you asked to avoid. Would you like to proceed with this path, or do you have any other instructions?} \\
\texttt{I09} & I have a meeting on the 4th floor. Could you go to the 4th floor and find the laptop, then bring it to the lobby on the B floor? \annline{$\rightarrow$ the building has no 4th floor (Floors A--C only).} & Clarification\newline Request & \textit{I couldn't find any laptops on the 4th floor. Could you please confirm if it's in a different location or provide more details?} \\
\texttt{I10} & I need five books for the book club meeting. Please bring all five books from the living room on the B floor to the kitchen. \annline{$\rightarrow$ only three books available in the specified source.} & Clarification\newline Request & \textit{Only three books were found in the living room: \texttt{book\_202}, \texttt{book\_203}, and \texttt{book\_204}. We couldn't locate the remaining two books. Do you have any additional instructions or should we proceed with just these three books?} \\
\end{xltabular}
}

%% file: checklist.tex
\section*{NeurIPS Paper Checklist}

\begin{enumerate}

\item {\bf Claims}
    \item[] Question: Do the main claims made in the abstract and introduction accurately reflect the paper's contributions and scope?
    \item[] Answer: \answerYes{}
    \item[] Justification: The abstract and Introduction state the main claims---an ontology-grounded world model aligned with the PDDL domain, a tool-using workflow for grounding and symbolic planning, and improved task success and token efficiency---and the method and experiments support them. The paper also makes the scope explicit: fully observable symbolic indoor environments under controlled benchmark evaluation (Abstract; Sections~1, 3, 4, and~6).

\item {\bf Limitations}
    \item[] Question: Does the paper discuss the limitations of the work performed by the authors?
    \item[] Answer: \answerYes{}
    \item[] Justification: Section~\ref{sec:conclusion} discusses the main limitations actually emphasized in the paper: fully observable symbolic environments and conclusions limited to controlled planning quality rather than end-to-end real-world autonomy. Section~\ref{sec:experimental-setup} further makes the controlled benchmark scope explicit and notes that the benchmark tasks are manually authored.

\item {\bf Theory assumptions and proofs}
    \item[] Question: For each theoretical result, does the paper provide the full set of assumptions and a complete (and correct) proof?
    \item[] Answer: \answerNA{}
    \item[] Justification: The paper presents no theorems, lemmas, or formal proofs; its contributions are methodological and empirical (Sections~3--5).

    \item {\bf Experimental result reproducibility}
    \item[] Question: Does the paper fully disclose all the information needed to reproduce the main experimental results of the paper to the extent that it affects the main claims and/or conclusions of the paper (regardless of whether the code and data are provided or not)?
    \item[] Answer: \answerYes{}
    \item[] Justification: The paper specifies the evaluation protocol, metrics, shared LLM and planner settings, scene-scale construction, and agent prompts needed to reproduce the main results (Sections~3--5; Appendix~\ref{app:agent-prompts}; Appendix~\ref{app:dataset-scale-augmentation}; Appendix~\ref{app:experimental-protocol-details}).

\item {\bf Open access to data and code}
    \item[] Question: Does the paper provide open access to the data and code, with sufficient instructions to faithfully reproduce the main experimental results, as described in supplemental material?
    \item[] Answer: \answerYes{}
    \item[] Justification: The code, benchmark tasks, environments, and evaluation scripts are released at the repository linked in the abstract, with installation and reproduction instructions. The paper and appendix additionally document the dataset construction, evaluation protocol, implementation settings, and prompt structure needed for faithful reproduction (Abstract; Section~\ref{sec:experimental-setup}; Appendix~\ref{app:agent-prompts}; Appendix~\ref{app:experimental-protocol-details}).

\item {\bf Experimental setting/details}
    \item[] Question: Does the paper specify all the training and test details (e.g., data splits, hyperparameters, how they were chosen, type of optimizer) necessary to understand the results?
    \item[] Answer: \answerYes{}
    \item[] Justification: Section~\ref{sec:experimental-setup} and Appendix~\ref{app:experimental-protocol-details} specify the environments, scene scales, task subsets, scoring checks, shared LLM setting, planner backend, and search configuration. Training-specific details such as optimizer choice are not applicable because the method is not trained; agent prompts are documented in Appendix~\ref{app:agent-prompts}.

\item {\bf Experiment statistical significance}
    \item[] Question: Does the paper report error bars suitably and correctly defined or other appropriate information about the statistical significance of the experiments?
    \item[] Answer: \answerYes{}
    \item[] Justification: The paper reports paired bootstrap 95\% confidence intervals for the main general-task comparisons and the ablations, together with exact McNemar tests on per-task Task Success for the ablations. Appendix~\ref{app:statistical-significance} describes the stratified paired bootstrap procedure, 10,000 resamples, and percentile-based intervals for the four main performance metrics.

\item {\bf Experiments compute resources}
    \item[] Question: For each experiment, does the paper provide sufficient information on the computer resources (type of compute workers, memory, time of execution) needed to reproduce the experiments?
    \item[] Answer: \answerYes{}
    \item[] Justification: All LLM calls are made through the OpenAI API, and the paper reports the token usage of every method (Sections~4--5). The world model (GraphDB) and the planner (Fast Downward) run locally on a single CPU workstation without a GPU (Appendix~\ref{app:experimental-protocol-details}).
    
\item {\bf Code of ethics}
    \item[] Question: Does the research conducted in the paper conform, in every respect, with the NeurIPS Code of Ethics \url{https://neurips.cc/public/EthicsGuidelines}?
    \item[] Answer: \answerYes{}
    \item[] Justification: The study uses symbolic indoor benchmarks, involves no human subjects or crowdsourcing, and does not release high-risk generative models or scraped datasets (Sections~4 and~6). To the best of our knowledge, it conforms to the NeurIPS Code of Ethics.

\item {\bf Broader impacts}
    \item[] Question: Does the paper discuss both potential positive societal impacts and negative societal impacts of the work performed?
    \item[] Answer: \answerNA{}
    \item[] Justification: The paper studies planning quality in fully observable symbolic indoor benchmark environments and confines evaluation to controlled symbolic experiments rather than real-world deployment or real-robot execution studies (Sections~4--6). We therefore treat direct societal impacts of the submitted work as not applicable.
    
\item {\bf Safeguards}
    \item[] Question: Does the paper describe safeguards that have been put in place for responsible release of data or models that have a high risk for misuse (e.g., pre-trained language models, image generators, or scraped datasets)?
    \item[] Answer: \answerNA{}
    \item[] Justification: The paper does not release high-risk generative models, scraped datasets, or similar assets. It studies a robot task-planning framework in symbolic indoor environments rather than releasing assets with elevated misuse risk (Sections~3--5).

\item {\bf Licenses for existing assets}
    \item[] Question: Are the creators or original owners of assets (e.g., code, data, models), used in the paper, properly credited and are the license and terms of use explicitly mentioned and properly respected?
    \item[] Answer: \answerYes{}
    \item[] Justification: The 3D Scene Graph dataset~\citep{armeni20193d} is used under its non-commercial research-only license, and the derived scene files we release are distributed under the same terms. Fast Downward~\citep{helmert2006fast} is used under GPLv3, and GPT-4o and the OpenAI embedding model are accessed through the OpenAI API under the OpenAI Services Agreement.

\item {\bf New assets}
    \item[] Question: Are new assets introduced in the paper well documented and is the documentation provided alongside the assets?
    \item[] Answer: \answerYes{}
    \item[] Justification: The released repository documents the benchmark tasks and their evaluation conditions, the environment files at the three scales, their construction, and the license terms of the source data, alongside the paper's description (Section~\ref{sec:experimental-setup}; Appendix~\ref{app:general-instruction-examples}; Appendix~\ref{app:dataset-scale-augmentation}).

\item {\bf Crowdsourcing and research with human subjects}
    \item[] Question: For crowdsourcing experiments and research with human subjects, does the paper include the full text of instructions given to participants and screenshots, if applicable, as well as details about compensation (if any)? 
    \item[] Answer: \answerNA{}
    \item[] Justification: The study uses authored benchmark tasks in symbolic indoor environments and includes no crowdsourcing or human-subject experiments (Section~\ref{sec:experimental-setup}).

\item {\bf Institutional review board (IRB) approvals or equivalent for research with human subjects}
    \item[] Question: Does the paper describe potential risks incurred by study participants, whether such risks were disclosed to the subjects, and whether Institutional Review Board (IRB) approvals (or an equivalent approval/review based on the requirements of your country or institution) were obtained?
    \item[] Answer: \answerNA{}
    \item[] Justification: No human participants or crowdsourced data were involved, so IRB approval or an equivalent review was not applicable (Section~\ref{sec:experimental-setup}).

\item {\bf Declaration of LLM usage}
    \item[] Question: Does the paper describe the usage of LLMs if it is an important, original, or non-standard component of the core methods in this research? Note that if the LLM is used only for writing, editing, or formatting purposes and does \emph{not} impact the core methodology, scientific rigor, or originality of the research, declaration is not required.
    \item[] Answer: \answerYes{}
    \item[] Justification: LLM usage is central to the method and is documented in the introduction, method, experimental setup, and appendix: GPT-4o-based agents interpret instructions, query tools, formalize constraints, and coordinate symbolic planning (Sections~1, 3, and~4; Appendix~\ref{app:agent-prompts}).

\end{enumerate}

%% file: references.bib
@inproceedings{ruan2023tptu,
  title={{TPTU}: Task Planning and Tool Usage of Large Language Model-based {AI} Agents},
  author={Ruan, Jingqing and Chen, Yihong and Zhang, Bin and Xu, Zhiwei and Bao, Tianpeng and Du, Guoqing and Shi, Shiwei and Mao, Hangyu and Li, Ziyue and Zeng, Xingyu and Zhao, Rui},
  booktitle={NeurIPS 2023 Foundation Models for Decision Making Workshop},
  year={2023}
}

@inproceedings{yang2024swe,
  title={{SWE}-agent: Agent-Computer Interfaces Enable Automated Software Engineering},
  author={Yang, John and Jimenez, Carlos E. and Wettig, Alexander and Lieret, Kilian and Yao, Shunyu and Narasimhan, Karthik and Press, Ofir},
  booktitle={Advances in Neural Information Processing Systems},
  volume={37},
  pages={50528--50652},
  year={2024}
}

@inproceedings{jiang2025artificial,
  title={Artificial Hivemind: The Open-Ended Homogeneity of Language Models (and Beyond)},
  author={Jiang, Liwei and Chai, Yuanjun and Li, Margaret and Liu, Mickel and Fok, Raymond and Dziri, Nouha and Tsvetkov, Yulia and Sap, Maarten and Choi, Yejin},
  booktitle={Advances in Neural Information Processing Systems},
  volume={38},
  year={2025}
}

@article{xie2023translating,
  title={Translating Natural Language to Planning Goals with Large-Language Models},
  author={Xie, Yaqi and Yu, Chen and Zhu, Tongyao and Bai, Jinbin and Gong, Ze and Soh, Harold},
  journal={arXiv preprint arXiv:2302.05128},
  year={2023}
}

@inproceedings{yang2024embodied,
  title={Embodied Multi-Modal Agent trained by an {LLM} from a Parallel {TextWorld}},
  author={Yang, Yijun and Zhou, Tianyi and Li, Kanxue and Tao, Dapeng and Li, Lusong and Shen, Li and He, Xiaodong and Jiang, Jing and Shi, Yuhui},
  booktitle={Proceedings of the IEEE/CVF Conference on Computer Vision and Pattern Recognition (CVPR)},
  pages={26275--26285},
  year={2024}
}

@inproceedings{weir2024learning,
  title={Learning to Reason via Program Generation, Emulation, and Search},
  author={Weir, Nathaniel and Khalifa, Muhammad and Qiu, Linlu and Weller, Orion and Clark, Peter},
  booktitle={Advances in Neural Information Processing Systems},
  volume={37},
  pages={36390--36413},
  year={2024}
}

@inproceedings{wang2023can,
  title={Can Language Models Solve Graph Problems in Natural Language?},
  author={Wang, Heng and Feng, Shangbin and He, Tianxing and Tan, Zhaoxuan and Han, Xiaochuang and Tsvetkov, Yulia},
  booktitle={Advances in Neural Information Processing Systems},
  volume={36},
  pages={30840--30861},
  year={2023}
}

@inproceedings{valmeekam2023planbench,
  title={{PlanBench}: An Extensible Benchmark for Evaluating Large Language Models on Planning and Reasoning about Change},
  author={Valmeekam, Karthik and Marquez, Matthew and Olmo, Alberto and Sreedharan, Sarath and Kambhampati, Subbarao},
  booktitle={Advances in Neural Information Processing Systems},
  volume={36},
  pages={38975--38987},
  year={2023}
}

@article{liu2023llm+,
  title={{LLM+P}: Empowering Large Language Models with Optimal Planning Proficiency},
  author={Liu, Bo and Jiang, Yuqian and Zhang, Xiaohan and Liu, Qiang and Zhang, Shiqi and Biswas, Joydeep and Stone, Peter},
  journal={arXiv preprint arXiv:2304.11477},
  year={2023}
}

@article{liu2024lost,
  title={{Lost in the Middle}: How Language Models Use Long Contexts},
  author={Liu, Nelson F. and Lin, Kevin and Hewitt, John and Paranjape, Ashwin and Bevilacqua, Michele and Petroni, Fabio and Liang, Percy},
  journal={Transactions of the Association for Computational Linguistics},
  volume={12},
  pages={157--173},
  year={2024}
}

@inproceedings{silver2022pddl,
  title={{PDDL} Planning with Pretrained Large Language Models},
  author={Silver, Tom and Hariprasad, Varun and Shuttleworth, Reece S. and Kumar, Nishanth and Lozano-P{\'e}rez, Tom{\'a}s and Kaelbling, Leslie Pack},
  booktitle={NeurIPS 2022 Foundation Models for Decision Making Workshop},
  year={2022}
}

@inproceedings{armeni20193d,
  title={{3D} Scene Graph: A Structure for Unified Semantics, {3D} Space, and Camera},
  author={Armeni, Iro and He, Zhi-Yang and Gwak, JunYoung and Zamir, Amir R. and Fischer, Martin and Malik, Jitendra and Savarese, Silvio},
  booktitle={Proceedings of the IEEE/CVF International Conference on Computer Vision (ICCV)},
  pages={5664--5673},
  year={2019}
}

@article{hughes2024foundations,
  title={Foundations of Spatial Perception for Robotics: Hierarchical Representations and Real-Time Systems},
  author={Hughes, Nathan and Chang, Yun and Hu, Siyi and Talak, Rajat and Abdulhai, Rumaia and Strader, Jared and Carlone, Luca},
  journal={The International Journal of Robotics Research},
  volume={43},
  number={10},
  pages={1457--1505},
  year={2024}
}

@inproceedings{wu2024can,
  title={Can Graph Learning Improve Planning in {LLM}-based Agents?},
  author={Wu, Xixi and Shen, Yifei and Shan, Caihua and Song, Kaitao and Wang, Siwei and Zhang, Bohang and Feng, Jiarui and Cheng, Hong and Chen, Wei and Xiong, Yun and Li, Dongsheng},
  booktitle={Advances in Neural Information Processing Systems},
  volume={37},
  pages={5338--5383},
  year={2024}
}

@inproceedings{kambhampati2024position,
  title={Position: {LLMs} Can't Plan, but Can Help Planning in {LLM}-Modulo Frameworks},
  author={Kambhampati, Subbarao and Valmeekam, Karthik and Guan, Lin and Verma, Mudit and Stechly, Kaya and Bhambri, Siddhant and Saldyt, Lucas Paul and Murthy, Anil B.},
  booktitle={Proceedings of the 41st International Conference on Machine Learning},
  series={Proceedings of Machine Learning Research},
  volume={235},
  pages={22895--22907},
  year={2024}
}

@inproceedings{valmeekam2023planning,
  title={On the Planning Abilities of Large Language Models: A Critical Investigation},
  author={Valmeekam, Karthik and Marquez, Matthew and Sreedharan, Sarath and Kambhampati, Subbarao},
  booktitle={Advances in Neural Information Processing Systems},
  volume={36},
  pages={75993--76005},
  year={2023}
}

@inproceedings{stechly2024chain,
  title={{Chain of Thoughtlessness?} An Analysis of {CoT} in Planning},
  author={Stechly, Kaya and Valmeekam, Karthik and Kambhampati, Subbarao},
  booktitle={Advances in Neural Information Processing Systems},
  volume={37},
  pages={29106--29141},
  year={2024}
}

@inproceedings{li2024embodied,
  title={Embodied Agent Interface: Benchmarking {LLMs} for Embodied Decision Making},
  author={Li, Manling and Zhao, Shiyu and Wang, Qineng and Wang, Kangrui and Zhou, Yu and Srivastava, Sanjana and Gokmen, Cem and Lee, Tony and Li, Li Erran and Zhang, Ruohan and Liu, Weiyu and Liang, Percy and Fei-Fei, Li and Mao, Jiayuan and Wu, Jiajun},
  booktitle={Advances in Neural Information Processing Systems},
  volume={37},
  pages={100428--100534},
  year={2024}
}

@inproceedings{guan2023leveraging,
  title={Leveraging Pre-trained Large Language Models to Construct and Utilize World Models for Model-based Task Planning},
  author={Guan, Lin and Valmeekam, Karthik and Sreedharan, Sarath and Kambhampati, Subbarao},
  booktitle={Advances in Neural Information Processing Systems},
  volume={36},
  pages={79081--79094},
  year={2023}
}

@inproceedings{liu2025delta,
  title={{DELTA}: Decomposed Efficient Long-Term Robot Task Planning using Large Language Models},
  author={Liu, Yuchen and Palmieri, Luigi and Koch, Sebastian and Georgievski, Ilche and Aiello, Marco},
  booktitle={2025 IEEE International Conference on Robotics and Automation (ICRA)},
  pages={10995--11001},
  year={2025}
}

@inproceedings{rana2023sayplan,
  title={{SayPlan}: Grounding Large Language Models using {3D} Scene Graphs for Scalable Robot Task Planning},
  author={Rana, Krishan and Haviland, Jesse and Garg, Sourav and Abou-Chakra, Jad and Reid, Ian and Suenderhauf, Niko},
  booktitle={Proceedings of the 7th Conference on Robot Learning},
  series={Proceedings of Machine Learning Research},
  volume={229},
  pages={23--72},
  year={2023},
  publisher={PMLR}
}

@inproceedings{huang2022inner,
  title={Inner Monologue: Embodied Reasoning through Planning with Language Models},
  author={Huang, Wenlong and Xia, Fei and Xiao, Ted and Chan, Harris and Liang, Jacky and Florence, Pete and Zeng, Andy and Tompson, Jonathan and Mordatch, Igor and Chebotar, Yevgen and Sermanet, Pierre and Jackson, Tomas and Brown, Noah and Luu, Linda and Levine, Sergey and Hausman, Karol and Ichter, Brian},
  booktitle={Proceedings of the 6th Conference on Robot Learning},
  series={Proceedings of Machine Learning Research},
  volume={205},
  pages={1769--1782},
  year={2023},
  publisher={PMLR}
}

@inproceedings{wang2023describe,
  title={Describe, Explain, Plan and Select: Interactive Planning with {LLMs} Enables Open-World Multi-Task Agents},
  author={Wang, Zihao and Cai, Shaofei and Chen, Guanzhou and Liu, Anji and Ma, Xiaojian and Liang, Yitao},
  booktitle={Advances in Neural Information Processing Systems},
  volume={36},
  pages={34153--34189},
  year={2023}
}

@inproceedings{liang2024introspective,
  title={Introspective Planning: Aligning Robots' Uncertainty with Inherent Task Ambiguity},
  author={Liang, Kaiqu and Zhang, Zixu and Fern{\'a}ndez Fisac, Jaime},
  booktitle={Advances in Neural Information Processing Systems},
  volume={37},
  pages={71998--72031},
  year={2024}
}

@article{helmert2006fast,
  title={{The Fast Downward Planning System}},
  author={Helmert, Malte},
  journal={Journal of Artificial Intelligence Research},
  volume={26},
  pages={191--246},
  year={2006}
}

@article{hurst2024gpt,
  title={{GPT-4o System Card}},
  author={Hurst, Aaron and Lerer, Adam and Goucher, Adam P. and others},
  journal={arXiv preprint arXiv:2410.21276},
  year={2024}
}

@techreport{aeronautiques1998pddl,
  title={{PDDL}---The Planning Domain Definition Language},
  author={McDermott, Drew and Ghallab, Malik and Howe, Adele E. and Knoblock, Craig A. and Ram, Ashwin and Veloso, Manuela and Weld, Daniel S. and Wilkins, David E.},
  institution={Yale Center for Computational Vision and Control},
  number={CVC TR-98-003/DCS TR-1165},
  year={1998}
}

@inproceedings{gu2024conceptgraphs,
  title={{ConceptGraphs}: Open-Vocabulary {3D} Scene Graphs for Perception and Planning},
  author={Gu, Qiao and Kuwajerwala, Ali and Morin, Sacha and Jatavallabhula, Krishna Murthy and Sen, Bipasha and Agarwal, Aditya and Rivera, Corban and Paul, William and Ellis, Kirsty and Chellappa, Rama and Gan, Chuang and de Melo, Celso Miguel and Tenenbaum, Joshua B. and Torralba, Antonio and Shkurti, Florian and Paull, Liam},
  booktitle={2024 IEEE International Conference on Robotics and Automation (ICRA)},
  pages={5021--5028},
  year={2024}
}

@inproceedings{fatemi2024talk,
  title={Talk like a Graph: Encoding Graphs for Large Language Models},
  author={Fatemi, Bahare and Halcrow, Jonathan and Perozzi, Bryan},
  booktitle={The Twelfth International Conference on Learning Representations},
  year={2024}
}

@inproceedings{levy2024same,
  title={Same Task, More Tokens: the Impact of Input Length on the Reasoning Performance of Large Language Models},
  author={Levy, Mosh and Jacoby, Alon and Goldberg, Yoav},
  booktitle={Proceedings of the 62nd Annual Meeting of the Association for Computational Linguistics (Volume 1: Long Papers)},
  pages={15339--15353},
  year={2024}
}

@article{planowl2025,
  title={{PlanOwl}: Automated {PDDL} Files Generation from {OWL} Ontologies and Visual Language Models},
  author={Adamik, Mark and Forte, Paolo},
  journal={Proceedings of the AAAI Symposium Series},
  volume={7},
  number={1},
  pages={634--643},
  year={2025}
}

@article{gestrin2024nl2plan,
  title={{NL2Plan}: Robust {LLM}-Driven Planning from Minimal Text Descriptions},
  author={Gestrin, Elliot and Kuhlmann, Marco and Seipp, Jendrik},
  journal={arXiv preprint arXiv:2405.04215},
  year={2024},
  note={Presented at the ICAPS 2024 HAXP Workshop}
}

@inproceedings{ren2023robots,
  title={Robots That Ask For Help: Uncertainty Alignment for Large Language Model Planners},
  author={Ren, Allen Z. and Dixit, Anushri and Bodrova, Alexandra and Singh, Sumeet and Tu, Stephen and Brown, Noah and Xu, Peng and Takayama, Leila and Xia, Fei and Varley, Jake and Xu, Zhenjia and Sadigh, Dorsa and Zeng, Andy and Majumdar, Anirudha},
  booktitle={Proceedings of the 7th Conference on Robot Learning},
  series={Proceedings of Machine Learning Research},
  volume={229},
  pages={661--682},
  year={2023},
  publisher={PMLR}
}

@inproceedings{he2024gretriever,
  title={{G-Retriever}: Retrieval-Augmented Generation for Textual Graph Understanding and Question Answering},
  author={He, Xiaoxin and Tian, Yijun and Sun, Yifei and Chawla, Nitesh V. and Laurent, Thomas and LeCun, Yann and Bresson, Xavier and Hooi, Bryan},
  booktitle={Advances in Neural Information Processing Systems},
  volume={37},
  pages={132876--132907},
  year={2024}
}

@inproceedings{zhao2023commonsense,
  title={Large Language Models as Commonsense Knowledge for Large-Scale Task Planning},
  author={Zhao, Zirui and Lee, Wee Sun and Hsu, David},
  booktitle={Advances in Neural Information Processing Systems},
  volume={36},
  pages={31967--31987},
  year={2023}
}

@inproceedings{sun2023adaplanner,
  title={{AdaPlanner}: Adaptive Planning from Feedback with Language Models},
  author={Sun, Haotian and Zhuang, Yuchen and Kong, Lingkai and Dai, Bo and Zhang, Chao},
  booktitle={Advances in Neural Information Processing Systems},
  volume={36},
  pages={58202--58245},
  year={2023}
}

@inproceedings{shinn2023reflexion,
  title={{Reflexion}: Language Agents with Verbal Reinforcement Learning},
  author={Shinn, Noah and Cassano, Federico and Gopinath, Ashwin and Narasimhan, Karthik and Yao, Shunyu},
  booktitle={Advances in Neural Information Processing Systems},
  volume={36},
  pages={8634--8652},
  year={2023}
}

@inproceedings{schick2023toolformer,
  title={{Toolformer}: Language Models Can Teach Themselves to Use Tools},
  author={Schick, Timo and Dwivedi-Yu, Jane and Dess{\`i}, Roberto and Raileanu, Roberta and Lomeli, Maria and Hambro, Eric and Zettlemoyer, Luke and Cancedda, Nicola and Scialom, Thomas},
  booktitle={Advances in Neural Information Processing Systems},
  volume={36},
  pages={68539--68551},
  year={2023}
}

@inproceedings{huang2023grounded,
  title={Grounded Decoding: Guiding Text Generation with Grounded Models for Embodied Agents},
  author={Huang, Wenlong and Xia, Fei and Shah, Dhruv and Driess, Danny and Zeng, Andy and Lu, Yao and Florence, Pete and Mordatch, Igor and Levine, Sergey and Hausman, Karol and Ichter, Brian},
  booktitle={Advances in Neural Information Processing Systems},
  volume={36},
  pages={59636--59661},
  year={2023}
}

@inproceedings{yao2023react,
    title={{ReAct}: Synergizing Reasoning and Acting in Language Models},
    author={Yao, Shunyu and Zhao, Jeffrey and Yu, Dian and Du, Nan and Shafran, Izhak and Narasimhan, Karthik and Cao, Yuan},
    booktitle={The Eleventh International Conference on Learning Representations},
    year={2023}
}

@article{lu2026novelty,
    title={Novelty Adaptation Through Hybrid Large Language Model ({LLM})-Symbolic Planning and {LLM}-guided Reinforcement Learning}, 
    author={Lu, Hong and Lorang, Pierrick and Duggan, Timothy R and Sinapov, Jivko and Scheutz, Matthias},
    journal={arXiv preprint arXiv:2603.11351},
    year={2026}
}

@inproceedings{hughes2022hydra,
  title={{Hydra: A Real-time Spatial Perception System for 3D Scene Graph Construction and Optimization}},
  author={Hughes, Nathan and Chang, Yun and Carlone, Luca},
  booktitle={Proceedings of Robotics: Science and Systems},
  address={New York City, NY, USA},
  month={June},
  year={2022},
  doi={10.15607/RSS.2022.XVIII.050}
}

@inproceedings{gorlo2026daaam,
  title={Describe Anything Anywhere at Any Moment},
  author={Gorlo, Nicolas and Schmid, Lukas and Carlone, Luca},
  booktitle={Proceedings of the IEEE/CVF Conference on Computer Vision and Pattern Recognition (CVPR)},
  pages={35002--35013},
  year={2026}
}

@inproceedings{agia2022taskography,
  title={Taskography: Evaluating Robot Task Planning over Large {3D} Scene Graphs},
  author={Agia, Christopher and Jatavallabhula, Krishna Murthy and Khodeir, Mohamed and Miksik, Ondrej and Vineet, Vibhav and Mukadam, Mustafa and Paull, Liam and Shkurti, Florian},
  booktitle={Proceedings of the 5th Conference on Robot Learning},
  series={Proceedings of Machine Learning Research},
  volume={164},
  pages={46--58},
  year={2022}
}

@inproceedings{beetz2018knowrob,
  title={{KnowRob} 2.0 -- A 2nd Generation Knowledge Processing Framework for Cognition-Enabled Robotic Agents},
  author={Beetz, Michael and Be{\ss}ler, Daniel and Haidu, Andrei and Pomarlan, Mihai and Bozcuo{\u{g}}lu, Asil Kaan and Bartels, Georg},
  booktitle={2018 IEEE International Conference on Robotics and Automation (ICRA)},
  pages={512--519},
  year={2018}
}

@article{honerkamp2024language,
  title={Language-Grounded Dynamic Scene Graphs for Interactive Object Search with Mobile Manipulation},
  author={Honerkamp, Daniel and B{\"u}chner, Martin and Despinoy, Fabien and Welschehold, Tim and Valada, Abhinav},
  journal={IEEE Robotics and Automation Letters},
  volume={9},
  number={10},
  pages={8298--8305},
  year={2024}
}

@inproceedings{liu2026nlpddl,
  title={Natural Language {PDDL} ({NL-PDDL}) for Open-world Goal-oriented Commonsense Regression Planning in Embodied {AI}},
  author={Liu, Xiaotian and Toroghi, Armin and Liang, Jiazhou and Courtis, David and Li, Ruiwen and Pesaranghader, Ali and Kim, Jaehong and Sadhu, Tanmana and Jeon, Hyejeong and Sanner, Scott},
  booktitle={The Fourteenth International Conference on Learning Representations},
  year={2026}
}
